%% file: paper-semantic-CDMAKE.tex
\documentclass[runningheads]{llncs}

\usepackage{dsbda-style}

\usepackage{amsmath, amsfonts}
\usepackage[english]{babel}
\usepackage{booktabs}
\usepackage{breqn}
\usepackage{caption}
\usepackage{enumitem}
\usepackage{graphicx}
\usepackage{hyperref}

\usepackage{multirow}
\usepackage{pdfpages}
\usepackage{tabularx}
\newcolumntype{R}{>{\raggedleft\arraybackslash}X}
\usepackage{xcolor}

\usepackage{cite}
\usepackage{amsmath,amssymb,amsfonts}
\usepackage{algorithmic}
\usepackage{graphicx}
\usepackage{textcomp}
\usepackage{xcolor}
\def\BibTeX{{\rm B\kern-.05em{\sc i\kern-.025em b}\kern-.08em
    T\kern-.1667em\lower.7ex\hbox{E}\kern-.125emX}}

\usepackage{dsbda-style}
\usepackage{booktabs}
\usepackage{threeparttable}

\usepackage{bm}
\usepackage{enumitem}
\setlist[itemize]{align=parleft,left=8pt..16pt}
\usepackage{amsmath}
\usepackage[ruled, noend]{algorithm2e}
\usepackage{tabularx}
\usepackage{multirow}
\usepackage{xcolor}
\usepackage{graphicx}
\usepackage{caption}
\usepackage{subcaption}
\usepackage{float}
\restylefloat{table}
\usepackage{listings}
\usepackage{pdfpages}
\usepackage{array}
\usepackage{dsbda-style}

\usepackage{mdframed}

\definecolor{code_background}{rgb}{0.94,0.94,0.94}
\mdfdefinestyle{CommandFrame}{
    linecolor=black,
    innertopmargin=4pt,
    innerbottommargin=2pt,
    innerrightmargin=2pt,
    innerleftmargin=2pt,
    backgroundcolor=code_background,
    userdefinedwidth=\linewidth,
}
\mdfdefinestyle{SmallCommandFrame}{
    linecolor=black,
    innertopmargin=2pt,
    innerbottommargin=2pt,
    innerrightmargin=2pt,
    innerleftmargin=2pt,
    backgroundcolor=code_background,
    userdefinedwidth=\linewidth,
}

\newcolumntype{L}{>{\raggedright\arraybackslash}X}
\newcolumntype{V}[1]{>{\raggedright\arraybackslash}m{#1}}
\newcolumntype{C}{>{\centering\arraybackslash}X}
\newcolumntype{N}[1]{>{\centering\arraybackslash}p{#1}}
\newcolumntype{M}[1]{>{\centering\arraybackslash}m{#1}}

\newcommand\YAMLcolonstyle{\ttfamily\color{red}\mdseries}
\newcommand\YAMLkeystyle{\ttfamily\color{black}\bfseries\small}
\newcommand\YAMLvaluestyle{\ttfamily\color{blue}\mdseries}

\makeatletter
\def\NAT@spacechar{~}
\makeatother

\makeatletter

\newcommand\language@yaml{yaml}

\expandafter\expandafter\expandafter\lstdefinelanguage
\expandafter{\language@yaml}
{
  keywords={true,false,null,y,n},
  keywordstyle=\color{darkgray}\bfseries,
  basicstyle=\YAMLkeystyle,                                 
  sensitive=false,
  comment=[l]{\#},
  morecomment=[s]{/*}{*/},
  commentstyle=\color{purple}\ttfamily,
  stringstyle=\YAMLvaluestyle\ttfamily,
  moredelim=[l][\color{orange}]{\&},
  moredelim=[l][\color{magenta}]{*},
  moredelim=**[il][\YAMLcolonstyle{:}\YAMLvaluestyle]{:},   
  morestring=[b]',
  morestring=[b]",
  literate =    {---}{{\ProcessThreeDashes}}3
                {>}{{\textcolor{red}\textgreater}}1     
                {|}{{\textcolor{red}\textbar}}1 
                {\ -\ }{{\mdseries\ -\ }}3,
}

\lst@AddToHook{EveryLine}{\ifx\lst@language\language@yaml\YAMLkeystyle\fi}
\makeatother

\newcommand\ProcessThreeDashes{\llap{\color{cyan}\mdseries-{-}-}}

\usepackage[nonumberlist, acronym]{glossaries}
\newacronym{BERT}{BERT}{Bidirectional Encoder Representations from Transformers}
\newacronym{CNN}{CNN}{Convolutional Neural Network}
\newacronym{DVC}{DVC}{Dense Video Captioning}
\newacronym{MLP}{MLP}{Multi-layer Perceptron}
\newacronym{MT}{MT}{End-to-End Dense Video Captioning with Masked Transformer}
\newacronym{NP}{NP}{Noun Phrase}
\newacronym{Open IE}{Open IE}{Open Information Extraction}
\newacronym{PDVC}{PDVC}{End-to-End Dense Video Captioning with Parallel Decoding}
\newacronym{POS Tag}{POS Tag}{Part-of-Speech Tag}
\newacronym{tIoU}{tIoU}{Temporal Intersection over Union}

\makeglossaries

\begin{document}
\title{Event and Entity Extraction from \\ Generated Video Captions}

\author{Johannes Scherer\inst{1} \and 
Deepayan Bhowmik\inst{2} \orcidID{0000-0003-1762-1578} \and
Ansgar Scherp\inst{1} \orcidID{0000-0002-2653-9245}}
\authorrunning{J. Scherer, D. Bhowmik, and A. Scherp}

\institute{Universität Ulm, Germany
\email{firstname.lastname@uni-ulm.de} \and
Newcastle University, UK \email{deepayan.bhowmik@newcastle.ac.uk}}

\maketitle              
\begin{abstract}
\extended{Incorporating metadata is essential to enable users and programs to efficiently search, organize, and re-use data. Semantic metadata is data that describes the meaning of data or its content, thus providing context for interpretation for both, machines and humans. To make the best use of data, in particular multimedia data, it needs to be tagged with relevant semantic information.}
Annotation of multimedia data by humans is time-consuming and costly, while reliable automatic generation of semantic metadata is a major challenge. We propose a framework to extract semantic metadata solely from automatically generated video captions. As metadata, we consider entities, the entities' properties, relations between entities, and the video category. Our framework combines automatic video captioning models with natural language processing (NLP) methods. We use state-of-the-art dense video captioning models with masked transformer (MT) and parallel decoding (PVDC) to generate captions for videos of the ActivityNet Captions dataset. We analyze the output of the video captioning models using NLP methods. We evaluate the performance of our framework for each metadata type, while varying the amount of information the video captioning model provides. Our experiments show that it is possible to extract high-quality entities, their properties, and relations between entities. In terms of categorizing a video based on generated captions, the results can be improved. We observe that the quality of the extracted information is mainly influenced by the dense video captioning model's capability to locate events in the video and to generate the event captions.

We provide the source code here: \newline
\url{https://github.com/josch14/semantic-metadata-extraction-from-videos}

\end{abstract}


\section{Introduction}
\label{sec:introduction}

\extended{With the already existing and constantly rising amount of multimedia data available in digital format, it is desirable to have a standardized description of this data in the form of metadata. Programs and its users use metadata to efficiently search, arrange, and repurpose digital multimedia like image, video, and audio data~\cite{ContextForSemanticMetadata}. One can distinguish between \textit{low-level} and \textit{high-level} metadata. For videos, low-level metadata describes characteristics like color, texture, or shape, while high-level metadata allows descriptions on a conceptual level. This can be the digital rights, content summaries, or related content~\cite{MultimediaMetadataStandards}. The term \textit{semantic metadata} refers to data that describes the "meaning" of data. That means, it describes the content of the data to which it refers, and therefore is high-level. For videos, this can be the occurring objects or a summary of what is seen in the video.}

\extended{Semantic metadata is utilized in many ways. It provides means for annotating multimedia content with textual information, which enables various applications such as search and retrieval in databases~\cite{MPEG7}. In contrast, low-level metadata is in most cases not the data that users are interested in. The crucial difference between low-level and semantic or high-level metadata is that low-level metadata can be acquired automatically without user interaction, whereas high-level metadata typically needs to be annotated manually by the user~\cite{MultimediaMetadataStandards}. }
\shortorextended{The}{However,} annotation of multimedia with semantic metadata by humans is time-consuming and costly. Automatic extraction methods exist for different types of high-level metadata, but these methods usually have high error rates and therefore manual correction of the user is still required~\cite{MetadataCreationSystem}. Thus, in contrast to the value of semantic metadata, especially when it can be generated automatically, the reliable automatic generation of semantic metadata is still a major challenge. For each semantic metadata type, one could use a different computer vision method to generate the data. For example, video object detection could be used to detect entities in a video, while video visual relation tagging methods find instances of relations between depicted entities. 
However, when using multiple methods, they need to be trained separately and the training is, especially for videos, computationally expensive. 

\shortorextended{

\begin{figure*}[ht]
 \includegraphics[width=1.2\linewidth]{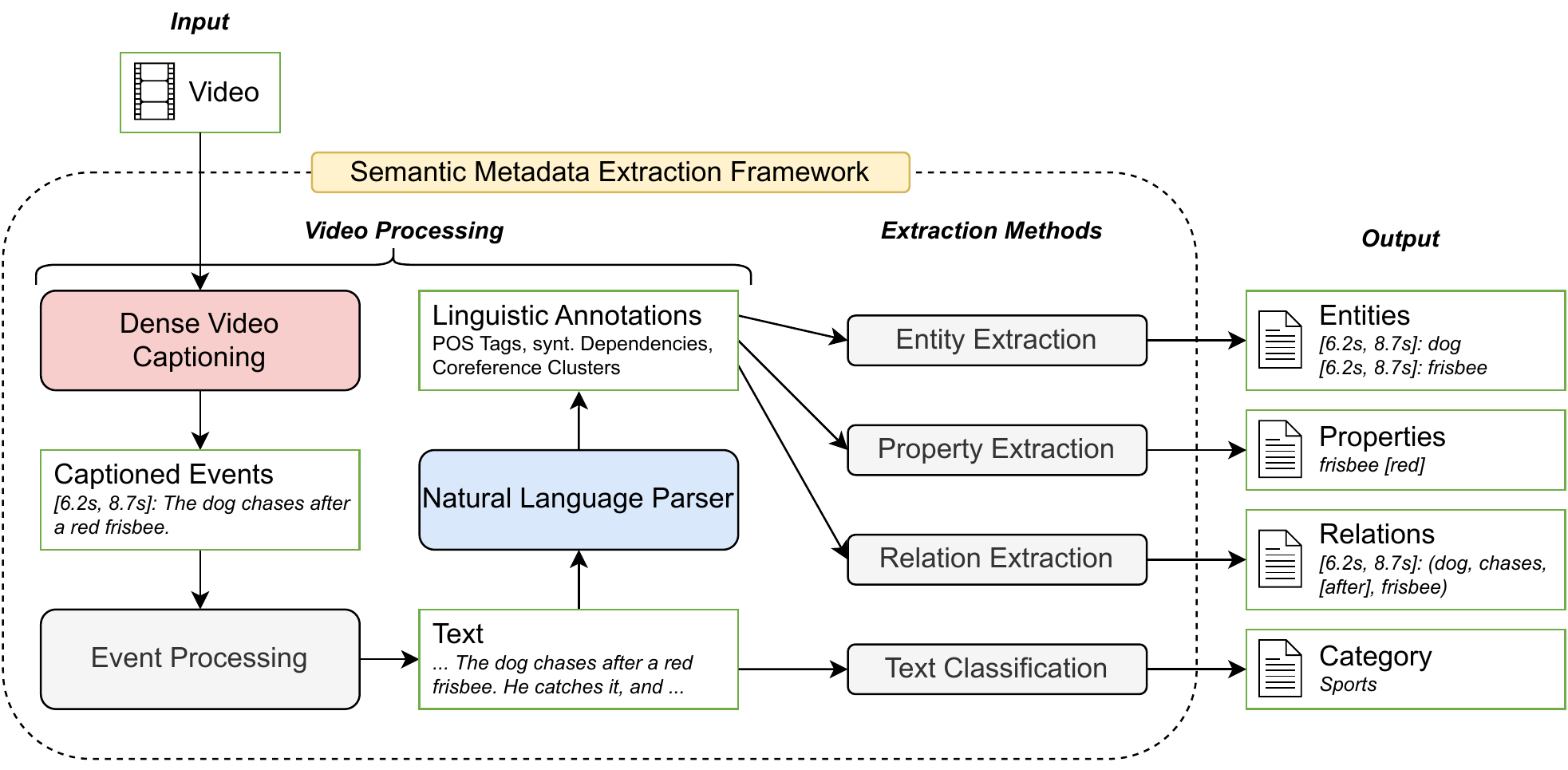}
\caption[Proposed framework for automatic semantic metadata extraction from videos.]{Semantic metadata extraction and its key components: a dense video captioning model and a natural language parser}
\label{fig:method_overview_1}
\end{figure*}

}{

 \begin{figure}[th]
     \centering
     \includegraphics[width=1.2\linewidth]{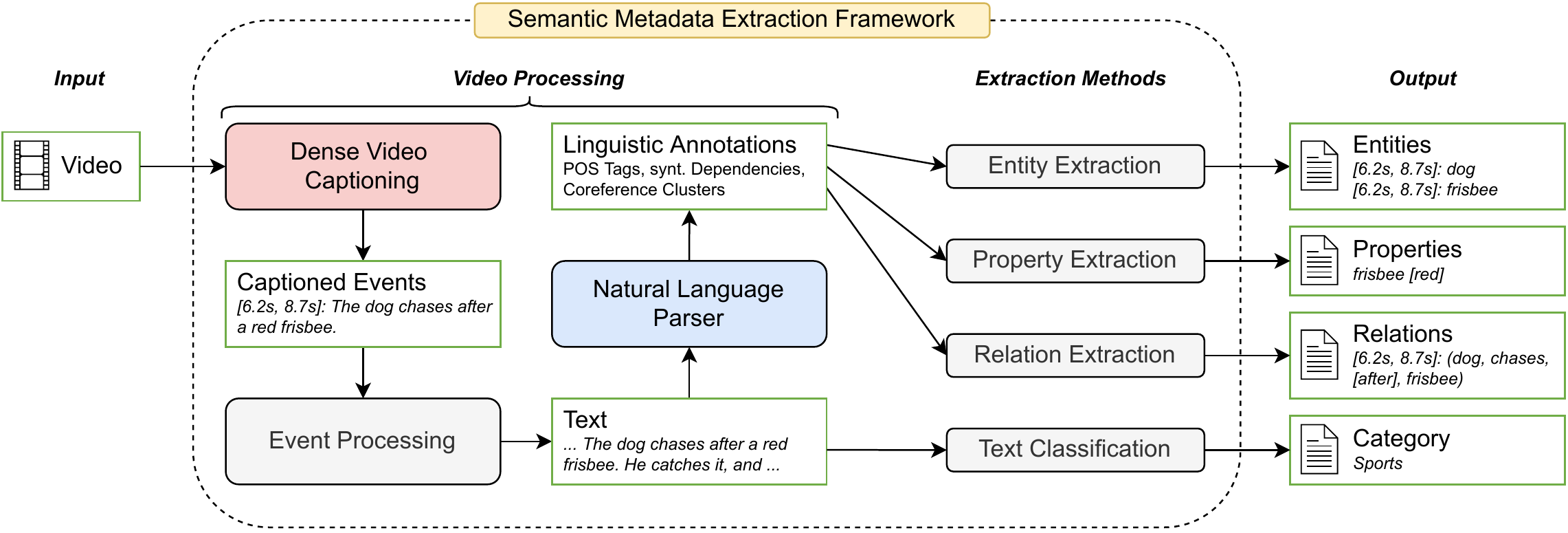}
     
     \caption{Semantic metadata extraction and its key components: a dense video captioning model and a natural language parser.   
     Our framework combines several computer vision and natural language processing methods. A dense video captioning model captures video semantics, which are processed into text. Furthermore, linguistic annotations are generated by a language parser. For each of the four semantic metadata types, a specific extraction method is used that exploits the processed video semantics.}
     \label{fig:method_overview_1_framework}
 \end{figure}

\begin{figure}[th]
\centering
    \includegraphics[width=.9\linewidth]{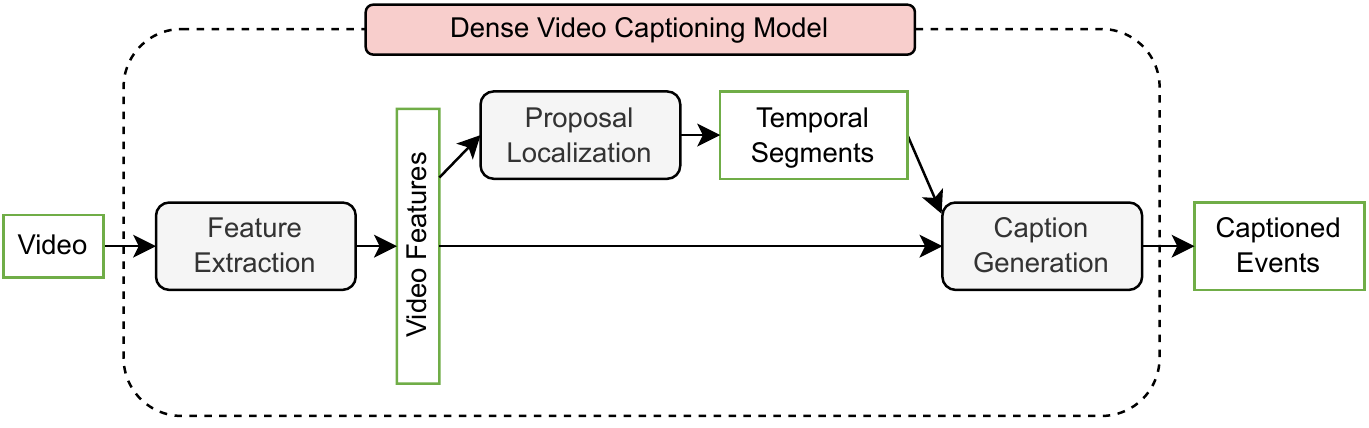}
    \caption{
      Dense video captioning models generate natural language sentences for multiple temporally localized video events, \ie captioned events. Such models often consist of three main components: the video encoder extracts visual features, the proposal decoder generates event proposals, and the captioning decoder generates a caption for each event proposal. Adapted from~\cite{PDVC}.
    }
    \label{fig:method_overview_1_DVC}
\end{figure}

\begin{figure}[th]
\centering
    \includegraphics[width=.8\linewidth]{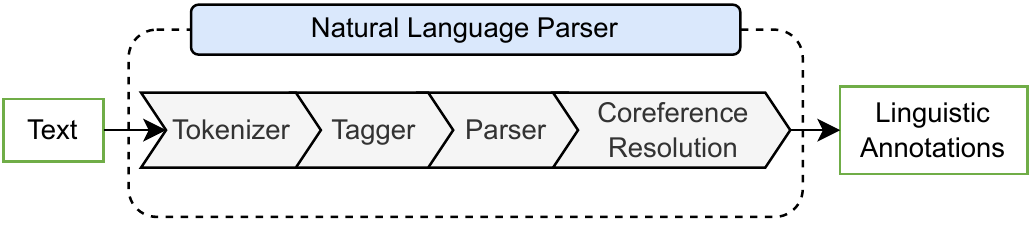}
    \caption{
        In a pipe-lined manner, language parser take in raw text and run a series of natural language annotators on the text. The semantic metadata extraction methods of our framework rely on the part-of-speech tags, syntactic dependencies, and coreference clusters generated by the language parser. \extended{Figure adapted from \url{https://spacy.io/usage/processing-pipelines}.}
   } 
    \label{fig:method_overview_1_NLP}
\end{figure}

}

From this motivation, we propose a framework that generates semantic metadata from videos of not only one, but multiple types. Depending on the video application, there are various semantic metadata types of interest. We focus on four different of those types, namely the depicted \textit{entities} and their \textit{properties}, the observable \textit{relations} between entities and the video \textit{category}. Additionally, we consider semantic metadata on different levels, namely event-level, where temporal information is relevant, and video-level. Our framework combines several methods from the fields of computer vision and natural language processing (NLP) (see Figures~\ref{fig:method_overview_1_framework} to \ref{fig:method_overview_1_NLP}). For an input video, a dense video captioning (DVC) model generates a set of natural language sentences for multiple temporally localized video events, thus providing a richly annotated description of video semantics. We process the captioned events into text to make them accessible for different NLP methods. 
Text classification determines the category of a video, while the extraction methods for entities, properties, and relations rely on linguistic annotations of a language parser. 
In summary, our contributions are:

\begin{itemize}
\item A framework for extracting semantic metadata combining an automatic video captioning model with several NLP methods for entity detection, extraction of entity properties, relation extraction, and categorical text classification.

\item We evaluate the capabilities of our framework \extended{to extract semantic metadata from automatically generated video captions} using the ActivityNet Captions~\cite{DVCPioneerANetCaptions} dataset. 
We compare two state-of-the-art dense video captioning models with masked transformer (MT)~\cite{MT} and parallel decoding (PVDC)~\cite{PDVC}.

\item The \extended{experiments show that the} quality of the extracted metadata mainly depends on \extended{the quality of~}the event localization in the video and the performance of the event caption generation. 

\end{itemize}

\extended{The remainder is organized as follow.}
Below, we discuss the related work\extended{ on semantic metadata modeling and extraction from dense video captioning}.
Section~\ref{sec:method} introduces the methods used to extract semantic metadata in the form of entities, properties, relations, and categories.
Section~\ref{sec:experiments} describes our experimental apparatus\extended{, including the datasets, procedure, hyperparameters, and evaluation}.
The results of our experiments are reported in Section~\ref{sec:results} and discussed in Section~\ref{sec:discussion}\extended{, before we conclude}.

\section{Related Work}
\label{sec:related_work}
\extended{Visual information of videos can be regarded at several abstraction levels, from low-level to semantic aspects of content. Video data models organize visual information to enable effective retrieval in video databases at different levels of abstractions~\cite{ApproachForModelingAndQueryingVideoData, DataModel}.
In Section~\ref{sec:modelling_semantic_metadata}, we present different video data models that include semantic information and show what types of semantic metadata they consider. In Section~\ref{sec:computer_vision_methods}, we present different computer vision methods, which in a way specialize in the extraction of a single type of semantic metadata.
Our extraction methods for the semantic metadata types entities, properties, and relations rely on the linguistic annotations produced by the language parser.
In Section~\ref{sec:open_information_extraction}, we present methods from the task of open information extraction that use a similar methodology. 
Finally, we present related work on text classification in Section~\ref{sec:text_classification_related_work}.}

\extended{
\subsection{Modelling Semantic Metadata for Videos}
\label{sec:modelling_semantic_metadata}
Jain and Hampapur~\cite{MetadataInVideoDatabases} presented a data model for videos based on a study of the applications of video, the nature of video retrieval requests, and the features of video. They argue that, unlike text, video data is opaque to computers. Therefore, to enable content-based access of video, a video database needs to possess some representation, \ie metadata in addition to the video. In their data model, the basic representational unit for videos are temporal segments. The authors distinguish between different metadata types. Meta features capture content-independent features of a video like the date of production. For content-dependent features, they distinguish between image features (based on a single frame) and video features (based on a time interval). Note that not all content-dependent features are semantic (\eg image brightness). Among others, the semantic metadata types that their data model considers are name, color, location and structure of objects, spacial activities, subjective properties like intentions and emotions, and the video category.

Due to the explosive growth in the number of online videos, Algur et al.~\cite{MetadataConstructionModelForWebVideos} stress the importance of descriptive metadata to enable efficient query-based video retrieval. They state that, for this purpose, the proper genre or category identification of a video is essential. Their proposed data model gives much more importance to descriptive metadata, for which they distinguish between three types. Among others, the video content summary is classified as general descriptive metadata. Event specific metadata enables users to extract specific video parts or events from videos. For an event, this includes its starting and end time, a descriptive event summary, and the event category (\eg discussion or fight). Lastly, object specific metadata stores information about objects like its caption, motion, shape, and color.

The data model of Hacid et al.~\cite{ApproachForModelingAndQueryingVideoData} for content-based video indexing and retrieval considers two types of information: 1)~entities that are described with attributes and modelled with relationships among them in the domain of a video sequence, and 2)~video frames which contain these entities. 
Their model explicitly links objects with relations to specify semantics of video data. 
They argue that, since information about video content is only true at a given time, temporal management of video information is required. For this, they discuss different approaches of how temporal segments are annotated to elements of interest. 
For example, each element can be associated with a single temporal descriptor. 
They extend this approach by defining temporal cohesion, which allows a set of non-overlapping time segments to be associated with a description. This allows the handling of all occurrences of an entity in a video document with a single object.

The presented video data models offer similar fields for the representation of semantic metadata, most importantly the video category, and entities or objects and their properties like color, location, and shape. Somewhat more specifically, the model of Jain and Hampapur considers subjective properties like emotions. The model of Algur et al. provides fields for video-level and event-level sentence descriptions of content, while the model of Hacid et al. model allows to link entities with relations. It is common to limit the validity of information to a certain interval.}
    
\subsection{Dense Video Captioning}
\label{sec:computer_vision_methods}
For each of the semantic metadata types entities, their properties, relations, and the video categories, one could think of a computer vision method to extract only a certain type. For example, video object detection involves object recognition, that means, identifying objects of different classes, and object tracking, \ie determining the position and size of an object in subsequent frames~\cite{VideoObjectDetectionTask}. Therefore, an object detection model could be used to determine the entities of a video and the information about when these are visible. 
Shang et al.~\cite{VidVRD} propose video visual relation tagging to detect relations between objects in videos. Here, relations are annotated to the whole video without the requirement of object localization. A relation is denoted by a triplet \textit{(subject, predicate, object)}, where the predicate may be a transitive or intransitive verb, comparative, or spatial predicate. \extended{With visual relation tagging, semantic metadata about the entities presented in a video and how they are related to each other can be generated. According to the authors, this effectively supports various visual relation-based applications such as video retrieval and visual question answering.}
Further methods that could be used for the extraction of video semantics include video classification for determining the category of a video~\cite{VideoClassification}, and emotion recognition, which aims to classify videos into basic emotions~\cite{VideoEmotionRecognition}. However, \extended{as mentioned above,~}it is not efficient to use one computer vision method at a time for the extraction of only one semantic metadata type. 

Automatic video description involves understanding and detection of different types of information like background scene, humans, objects, human actions, and events like human-object interactions~\cite{SurveyVideoCaptioning1}. In such a way, automatic video description can be seen as a task that unites the mentioned computer vision tasks like object detection, visual relation tagging, and emotion recognition. Dense video captioning (DVC), as first introduced by Krishna et al.~\cite{DVCPioneerANetCaptions}, generate \textit{captioned events}, which not only involves the localization of multiple, potentially overlapping events in time, but also the generation of a natural language sentence description for each event. Because of the rich information DVC models provide, we utilize such model in our framework. 
We present two DVC models with masked transformer (MT)~\cite{MT} and parallel decoding (PVDC)~\cite{PDVC}, which we use in our experiments, in detail in Section~\ref{sec:method_dense_video_captioning_model} (together with our framework).

\subsection{Text Information Extraction and Classification}
\label{sec:open_information_extraction}

In our framework, semantic metadata is extracted from the captioned events generated by a DVC model. This includes the analysis of the events' textual descriptions, for which methods from Open Information Extraction (Open IE) can be employed~\cite{OpenIESurvey}. 
Open IE is the task of generating a structured representation of the information extracted from a natural language text in the form of relational triples. A triple \textit{(arg1, rel, arg2)} consists of a set of argument phrases and a phrase denoting a semantic relation between them~\cite{OpenIESurvey}.
Existing Open IE approaches make use of a set of patterns, which are either hand-crafted rules or automatically learned from labeled training data. Furthermore, both methodologies can be divided into two subcategories: approaches that use shallow syntactic analysis and approaches that utilize dependency parsing~\cite{OpenIESurvey2}.
Fader et al.~\cite{REVERB} proposed
R{\small E}V{\small ERB}, which makes use of hand-crafted extraction rules. They restrict syntactic analysis to part-of-speech tagging
and noun phrase chunking, resulting in an efficient extraction for high-confidence propositions. Relations are extracted in two major steps: first, relation phrases are identified that meet syntactic and lexical constraints. 
Then, for each relation phrase, a pair of noun phrase arguments is identified. 
Contrary to R{\small E}V{\small ERB}, ClausIE~(clause-based Open IE) uses hand-crafted extraction rules based on a typed dependency structure~\cite{ClausIE}.
It does not make use of any training data and does not require any postprocessing like filtering out low-precision extractions. 
First, a dependency parse of the sentence is computed. Then, using the dependency parse, a set of clauses is determined. The authors define seven clause types, where each clause consists of one subject, one verb and optionally of an indirect object, a direct object, a complement, and one or more adverbials. 
Finally, for each clause, one or more propositions are generated.
Since dependency parsing is used, ClausIE is computationally more expensive compared to R{\small E}V{\small ERB}. 
\extended{However, it achieves improved precision and recall.}

Algur et al.~\cite{MetadataConstructionModelForWebVideos} argue that the proper category identification of a video is essential for efficient query-based video retrieval.
\label{sec:text_classification_related_work}
This task is traditionally posed as a supervised classification of the features derived from a video. The features used for video classification can be of visual nature only, but if user-provided textual metadata (\ie title, description, tags) is available, it can be used in a profitable way~\cite{MetadataForClassification}. However, in our proposed framework the video category is predicted only with the textual information the DVC model provides. So although we address the video classification problem, we do this by utilizing existing work in the text classification area. 
Text classification models can be roughly divided into two categories~\cite{TextClassificationSurvey1}. 
First, traditional statistics-based models such as k-Nearest Neighbors and Support Vector Machines require manual feature engineering. 
Second, deep learning models consist of artificial neural networks to automatically learn high-level features for better results in text understanding.
For example, TextCNN~\cite{TextCNN} is a text classification method using text-induced word-document cooccurence graph and graph learning.
We use the pre-trained BERT (Bidirectional Encoder Representations from Transformers)~\cite{BERT} model, which set the state-of-the-art for text classification~\cite{DBLP:conf/acl/GalkeS22}.

\section{Semantic Metadata Extraction from Videos}
\label{sec:method}

\begin{figure*}[ht]
\centering
\includegraphics[width=1.2\textwidth]{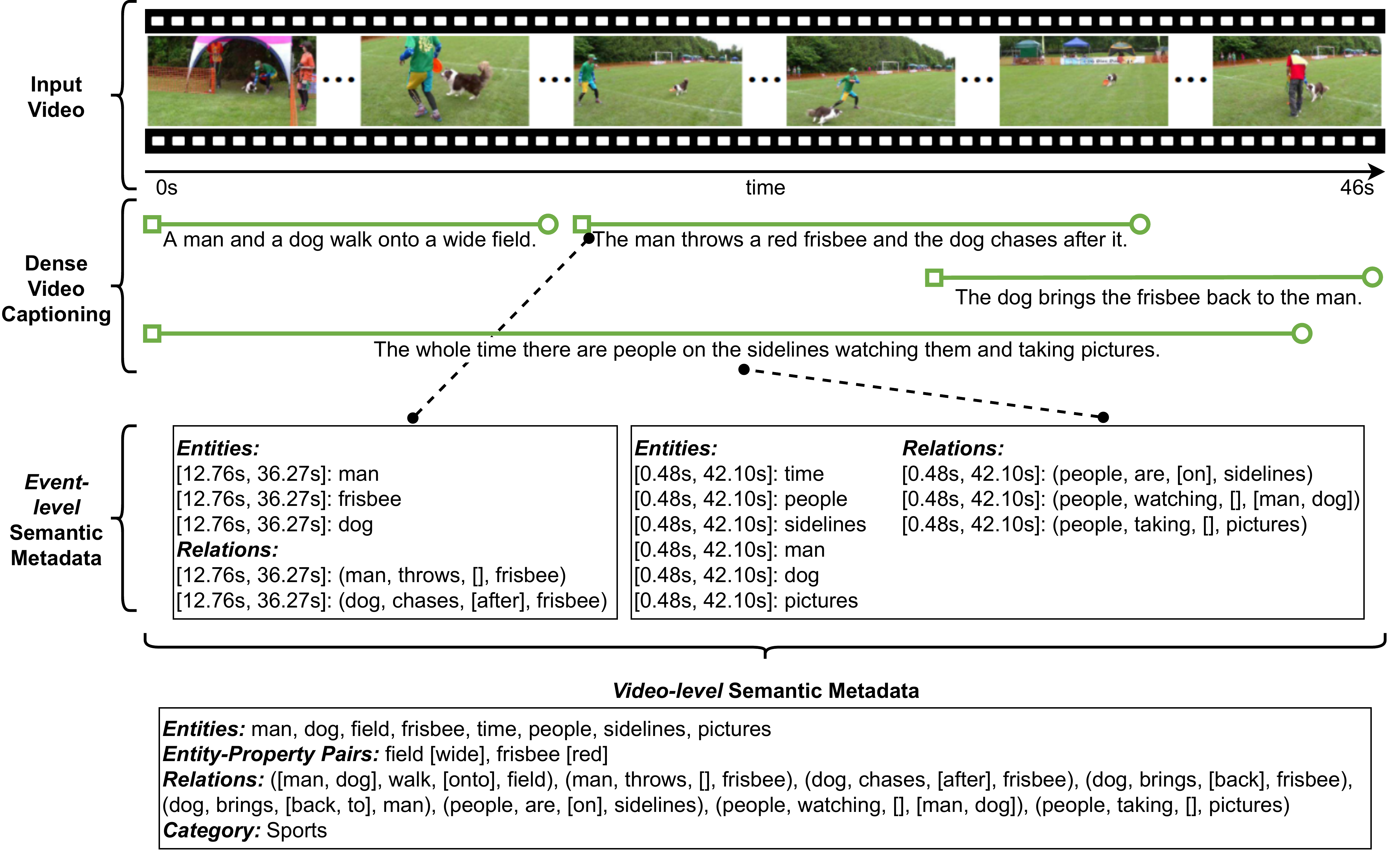}
\caption{Our framework extracts semantic metadata in the form of entities, properties of entities, relations between entities, and the video category from automatically generated captioned events. We distinguish event-level and video-level semantic metadata, depending on whether semantic metadata is assigned to a specific time interval or not. Image adapted from~\cite{DVCImageSource}.}
\label{fig:method_overview_2}
\end{figure*}

\extended{Section~\ref{sec:modelling_semantic_metadata} gives an overview over the various types of semantic metadata that video data models consider.} 
In our framework, the extraction of semantic metadata is based only on the captioned events generated by the DVC model. As a result, certain metadata types such as emotions are difficult to extract. This depends mainly on how detailed the DVC model is able to describe video semantics.
Considering the capabilities of current video description models, we focus on four semantic metadata types that we aim to extract from a video: the depicted \textbf{entities} (\ie persons, objects, locations) and their \textbf{properties}, observable visual \textbf{relations} between entities, and the video \textbf{category}, see Figure~\ref{fig:method_overview_2}. 
We distinguish between \textit{event-level} and \textit{video-level} semantic metadata depending on whether semantic metadata is assigned to a specific time interval or not. For example, assume that at some point in a video there is a \textit{cat} visible. On video-level, the corresponding entity item only stores the information that there is a cat occurring in the video. On event-level, the metadata item does not only store the name of the entity, but also a time interval in which the entity is visible. 
\extended{Ultimately, this means that event-level semantic metadata refers to exactly one captioned event generated by the DVC model. 
In contrary, video-level semantic metadata is extracted from information provided by all video events.}

Revisiting Figure~\ref{fig:method_overview_1_framework}, it can be seen that our framework consists of several methods. For an input video, a DVC model generates captioned events (Sec.~\ref{sec:method_dense_video_captioning_model}), which are then processed into text to make them accessible for different methods (Sec.~\ref{sec:method_event_processing}). 
The natural language parser produces linguistic annotations, namely part-of-speech (POS) tags, a dependency parse, and coreference clusters (Sec.~\ref{sec:method_language_processing}). The extraction methods for the semantic metadata types entities (Sec.~\ref{sec:method_entity_extraction}), properties (Sec.~\ref{sec:method_property_extraction}), and relations (Sec.~\ref{sec:method_relation_extraction}) use these linguistic annotations, while the text classification method determines the category of a video using only the generated captioned events (Sec.~\ref{sec:method_text_classification}). The lexical database WordNet~\cite{WordNet} is used at various points to ensure that extracted semantic metadata consists of linguistically correct English nouns, verbs, adjectives, and adverbs.

\subsection{Dense Video Captioning (DVC)}
\label{sec:method_dense_video_captioning_model}
\extended{Dense video captioning is a multi-task problem that involves detecting multiple individual events in a video and understanding of their context. }
From an input video, DVC models generate a set of captioned events. Each captioned event consists of the event itself, a temporal segment which potentially overlaps with segments of other captioned events, and a natural language sentence that captions the event. While introducing the task of DVC, Krishna et al.~\cite{DVCPioneerANetCaptions} proposed a model which consists of a proposal module for event localization, and a separate captioning module, an attention-based Long Short-Term Memory network for context-aware caption generation.  
Zhou et al.~\cite{MT} argue that the model of Krishna et al. is not able to take advantage of language to benefit the event proposal module. To this end, they proposed an end-to-end DVC model with masked transformer (MT) that is able to simultaneously produce event proposals and event descriptions. Like many methods that tackle the DVC task, Zhou et al.'s model consists of three components\extended{~(see Figure \ref{fig:method_overview_1_DVC})}. 
The video encoder, composed of multiple self-attention layers, extracts visual features from video frames. 
The proposal decoder takes the features from the encoder and produces event proposals, \ie temporal segments. 
The captioning decoder takes input from \extended{both,~}the visual encoder and the proposal decoder\extended{,} to caption each event.

Wang et al.~\cite{PDVC} state that methods like the model of Zhou et al. follow a two-stage “localize-then-describe” scheme, which heavily relies on hand-crafted components. 
In contrast to the usual structure of DVC models\extended{ depicted in Figure~\ref{fig:method_overview_1_DVC}}, they proposed a simpler framework for end-to-end DVC with parallel decoding (PDVC). Their model directly decodes extracted frame features into a captioned event set by applying two parallel prediction heads: localization head and captioning head. 
They propose an event counter, which is stacked on top of the decoder to predict the number of final events. 
The authors claim that PVDC is able to precisely segment the video into a number of events, avoiding to miss semantic information as well as avoiding replicated caption generation.

\extended{We implement in our framework for semantic metadata extraction from videos  the two described models: MT and PDVC. 
To summarize, within our framework, the DVC model is responsible for generating captioned events, each consisting of a temporal segment and a sentence describing the event, which are then forwarded to the event processing module.}

\subsection{Event Processing}
\label{sec:method_event_processing}
\begin{figure}
\centering
\includegraphics[width=1\linewidth]{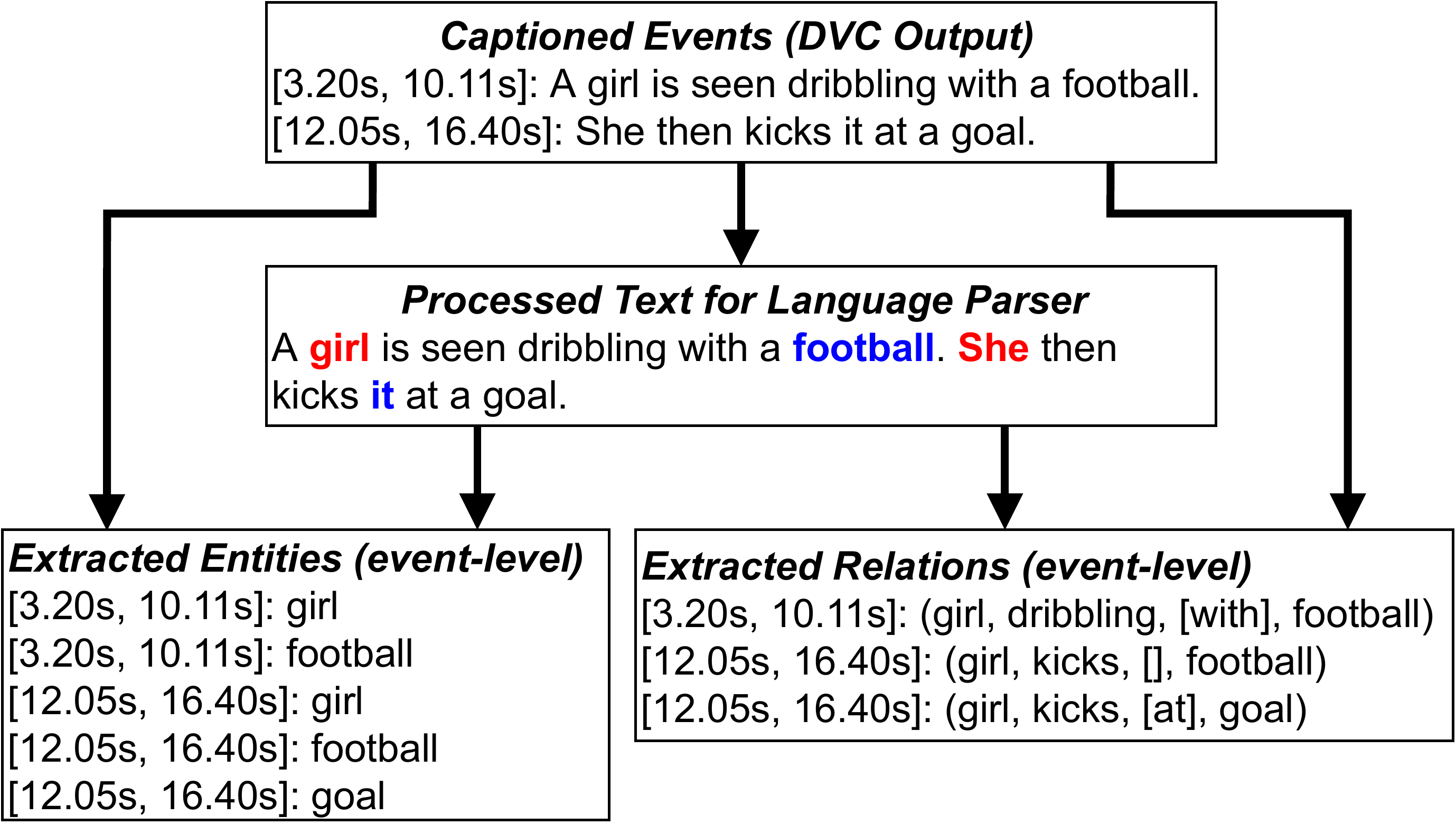}
\caption[Illustration of the processing of captioned events into text.]{Captioned events are processed into text and made accessible to the language parser and text classifier.
\extended{This also allows the language parser to use coreference resolution.}
}
\label{fig:event_processing}
\end{figure}

The event processing module processes the captioned events generated by the preceding DVC model into text in order to make semantic information accessible to the natural language parser and the text classification method. In detail, the sentences of the captioned events are sorted in ascending order of the start times of the corresponding events. Afterwards, the sentences are concatenated, resulting in a single text per video, and forwarded to the language parser and text classification method, respectively. By not passing the sentences separately to the language parser, this enables it to use coreference resolution (see Section~\ref{sec:method_language_processing}). When extracting entities and relations on event-level, we annotate each entity and relation with the temporal segment of the captioned event whose sentence contains the name of the entity or the words of the relation, resp. (see Figure~\ref{fig:event_processing}).

\subsection{Language Processing}
\label{sec:method_language_processing}
The extraction of entities, the entities' properties, and relations is done through syntactic analysis based on the linguistic annotations generated by the natural language parser. 
\shortorextended{It~}{More precisely, the language parser~}is required to provide POS tags of tokens, a dependency parse, and coreference clusters. 
The POS tag is a label assigned to the token to indicate its part of speech. 
The dependency parse consists of a set of directed syntactic relations between the tokens of the sentence. 
Coreference clusters \extended{are the result of coreference resolution, which }aim to find all language expressions that refer to the same entity in a text.

We use a CNN-based model from spaCy\footnote{\label{footnote:spaCy}spaCy is available at: \url{https://github.com/explosion/spaCy}.}\extended{, an open-source  library (for details on implementation see Section~\ref{sec:experiments})}.
The model generates the desired linguistic annotations in a pipe-lined manner\extended{ (see Fig.~\ref{fig:method_overview_1_NLP})}: 
First, the tokenizer segments the input text into tokens. 
Afterwards, the tagger and dependency parser assign POS tags and dependency labels to tokens, respectively. 
Finally, coreference clusters are determined using spaCy's NeuralCoref\extended{\footnote{\label{footnote:NeuralCoref}NeuralCoref is available at: \url{https://github.com/huggingface/neuralcoref}.}} extension. 
For an input sentence, the dependency parse produced by spaCy models is a tree where the head of a sentence, which is usually a verb, has no dependency. Every other token of the dependency tree has a dependency label that indicates its syntactic relation to its \textit{parent}. The \textit{children} of a token are all its immediate syntactic dependents, \ie the tokens of the dependency tree for which it is the parent. 
\extended{NeuralCoref uses a neural network, which adapts the scoring model of Clark and Manning~\cite{CoreferenceResolution}, to rank and determine pairs of mentions. The resulting coreference clusters are sets, where each set contains all mentions that are co-referring to the same entity.}

\extended{In the following sections, when explaining how the semantic metadata extraction methods use linguistic annotations, we use the Universal Dependencies POS tags\footnote{\label{footnote:UniversalDependencies}Universal Dependencies POS tags: \url{https://universaldependencies.org/u/pos/}.}~\cite{UniversalDependencies}. For dependencies however, we use the ClearNLP dependency labels\footnote{\label{footnote:clearNLP}ClearNLP dependency labels: \url{https://github.com/clir/clearnlp-guidelines/blob/master/md/specifications/dependency_labels.md}.} due to the fact that spaCy makes use of them. Note that the ClearNLP dependency labels are slightly different from the syntactic relations of Universal Dependencies.}
\extended{Table~\ref{tab:universal_dependencies_tags} and Table~\ref{tab:clear_nlp_dependency_labels} in the appendix list and explain the POS tags and dependency labels that are used in this work.}

\subsection{Entity Extraction}
\label{sec:method_entity_extraction}
The entity extraction method determines which (video-level) and when (event-level) entities like persons, objects, and locations are visible in the video. A \underline{video-level} \textbf{entity} only consists of a \textit{name} describing the entity. On \underline{event-level}, the entity additionally consists of a \textit{temporal segment} containing the information about when the entity is seen. Entities are extracted by determining (compound) nouns with the POS tags and dependency labels of tokens, and coreference clusters, which are all provided by the language parser.

\extended{First, compound nouns are detected, \ie nouns that consist of more than one word token. Candidates for compound nouns are lists of tokens. Such list is formed for a token that is not marked with the \texttt{compound} dependency, when it has a child (see Section~\ref{sec:method_language_processing}) marked as \texttt{compound}. The candidate list then consists of the token itself and all recursively found children that are marked as \texttt{compound}. Now, each candidate list is checked whether it consists of a compound noun. For this to hold, the candidate list needs to have at least one token that is a \texttt{NOUN} or \texttt{PROPN}. While using the tokens in order of occurrence in the corresponding sentence, it is checked if the concatenation of the words using either \textit{"\_"}, \textit{"-"} or an empty string is an open, hyphenated, or regular compound noun, respectively. This is the case when the formed word or its lemma is a WordNet noun. If no compound noun is found, then the procedure repeats with the two sublists that do not contain the first or the last token of the candidate list. This recursive procedure ends either when a compound noun is found or no compound noun can be formed anymore. In such way, not only \texttt{NOUN}+\texttt{NOUN} (\eg \textit{living\_room}) compounds are found, but also other combinations like \texttt{ADJ}+\texttt{NOUN} (\textit{plastic\_bags}), \texttt{NOUN}+\texttt{VERB} (\textit{mountain\_climbing}), or \texttt{PROPN}+\texttt{PROPN}+\texttt{PROPN} (\textit{new\_york\_city}). The determination of the remaining nouns, \ie nouns consisting of one token, is simple. For this, only tokens which are not part of a detected compound noun are used. Such token is a noun, if the language parser marked it as a \texttt{NOUN} or \texttt{PROPN}, and WordNet recognizes the token or its lemma as noun.

On video-level, the method is finished already. Here, for each noun from the set of all detected nouns an entity is created with the noun as the entity's name.} 
On event-level, before assigning a temporal segment to each name in order to obtain the entities, each pronoun is analysed whether it refers to a noun or not (see Figure~\ref{fig:event_processing}). 
First, the tokens of the text are filtered for pronouns\extended{, \ie for tokens having the \texttt{PRON} tag}. 
Using the computed coreference clusters, for each pronoun it is checked whether it refers to a noun that has been determined in the previous step or not. If this is the case, then the pronoun is replaced with the corresponding noun, \ie the name of a new entity. Finally, the event-level entities are built by assigning temporal segments to the names of the entities. Here, for an entity, the temporal segment is the segment of the corresponding sentence in which its name (or the pronoun that was previously replaced) occurs.

\subsection{Property Extraction}
\label{sec:method_property_extraction}
The property extraction method determines properties of entities such as their color, size, and shape. The method is only used to extract video-level information in order to collect information for an entity from different captioned events. An \textbf{entity-property pair} is a tuple consisting of an \textit{entity} (\ie its name) and a \textit{property}, which further describes the entity.
\extended{Coreference clusters are unimportant for the property extraction method. Therefore, only the tokens' POS tags and dependencies, computed by the language parser, are utilized here.} 
For each (video-level) entity, extracted in the previous section, properties are determined as follows. 
The candidate properties for an entity are the children of the corresponding token (or tokens for compound nouns) that are marked with the dependency labels.
We use WordNet to recognize the token or its lemma as an adjective.
Such candidate property is considered as a property of the entity. 
An entity-property pair is formed with the name of the entity and the detected property. 
In such way, the method results in a set of video-level entity-property pairs. There is no restriction on the tag a property token needs to have. Therefore, properties may be marked by the language parser as \texttt{ADJ}ective (\eg \textit{round} ball), \texttt{VERB} (\eg \textit{provoking} film), or other tags.

\subsection{Relation Extraction}
\label{sec:method_relation_extraction}
As seen for visual relation tagging (Section~\ref{sec:computer_vision_methods}) or Open IE (Section~\ref{sec:open_information_extraction}), relations are usually formulated as triples. In contrast, we define relations as follows. A \underline{video-level} \textbf{relation} is a 4-tuple of the form \textit{(subjects, verb, modifiers, objects)}. With a \textit{temporal segment}, an \underline{event-level} relation has an additional element containing the information about when the relation is observed. The first element \textit{subjects} is a list containing the names of the relation's acting entities. \textit{Objects}, also a list of names of entities, contains the entities that are the receiver of the action. Note that usually subjects and objects contain only a single entity. In some cases, such as in the sentence \textit{"A boy and a girl are seen playing football."}, multiple entities are the actors in a relation: \textit{([boy, girl], play, [], football)}. 
With the above definition, we are able to capture relations with different types of verbs: single-word verbs and multi-word verbs, \ie prepositional verbs (verb+preposition), phrasal verbs (verb+particle), and phrasal-prepositional verbs. In both cases, \textit{verb} contains the verb. \textit{Modifiers} is an empty list for single-word verbs. For multi-word verbs, however, \textit{modifiers} contains the verb's particles and prepositions. Both, particles and prepositions, provide information about how the verb and the object are related to each other. For example, the relation \textit{(girl, catches, [up, with], kids)} for the sentence \textit{"The girl catches up with the other kids."} is better understood than the relation \textit{(girl, catches, kids)}.

As for entity extraction, the relation extraction method utilizes POS tags and dependency labels of tokens and coreference clusters. In fact, the entity extraction method is used here in order to determine valid subjects and objects for relations. The relation extraction method proceeds in three steps: 
search for candidate verbs, 
search for candidate tuples consisting of a subject and a verb, and
search for corresponding objects and modifiers for each verb in a candidate tuple. 
\shortorextended{In brief, these steps are conducted by analyzing the dependency tree, exploiting the POS tags, and exploiting WordNet.
}{

\textbf{Search for Candidate Verbs.}
Candidate verbs are determined using Algorithm~\ref{alg:find_verbs}. For the above definition of relations to make sense, passive verbs are ignored. Verb tokens are tagged either as \texttt{VERB} or \texttt{AUX} (1), whereby passive verbs have at least one child token that has either the \texttt{nsubjpass}, \texttt{auxpass}, or \texttt{agent} dependency (2). The candidate verbs are obtained using those characteristics. Also, a candidate verb or its lemma needs to be included in WordNet to be considered as such (3).

\begin{algorithm}[ht]
\scriptsize
\caption{\textit{Determine non-passive verbs of a token list.}}
\label{alg:find_verbs}
\LinesNumbered
\DontPrintSemicolon
\SetKwInput{Input}{Input}
\SetKwInput{Output}{Output}
\SetKwInput{Constants}{Constants}
\SetKwProg{Function}{Function}{:}{}

\SetKwFunction{findverbs}{find\_verbs}
\SetKwFunction{iswordnetverb}{is\_wordnet\_verb}
\SetKwFunction{dep}{dep}
\SetKwFunction{tag}{tag}
\SetKwFunction{childtokens}{child\_tokens}
\SetKwFunction{append}{append}
\SetKwFunction{parenttoken}{parent\_token}

\Input{Text tokens $T = [t_1,\dots]$}
\Output{List $V = [v_1,\dots]$ of verbs $v_i$}
\Constants{$Tags$ = [\texttt{VERB}, \texttt{AUX}] (indicate verbs),\newline
$Deps$ = [\texttt{nsubjpass}, \texttt{auxpass}, \texttt{agent}] (indicate passive verbs)
}
\Function{\findverbs{$T$}}{
$V = [\ ]$\\
\For{$t$ \textbf{\upshape in} $T$}{
\textcolor{blue}{\tcp*[h]{(1) Search for verbs}}\\
    \If{
        \tag{$t$} $\in Tags$}{
\textcolor{blue}{\tcp*[h]{(2) Ignore passive verbs}}\\
            \For{$c \in$ \childtokens{$t$}}{
                \lIf{\dep{$c$} $\in Deps$\\}{\textbf{\upshape continue} with next $t$}
            
            }
\textcolor{blue}{\tcp*[h]{(3) Verb included in WordNet?}}\\
            \If{\iswordnetverb{$t$}}{$V.$\append{$t$}}
    }
}
\Return{$V$}
}
\end{algorithm}

\textbf{Search for Candidate Tuples}
For each candidate verb, a subject is determined using Algorithm~\ref{alg:find_subject}, resulting in candidate tuples. At this stage, a subject can either be an entity or pronoun. Using coreference resolution, a pronoun can later be resolved into an entity, thus creating a valid relation. A distinction is made between three different cases of how subjects can be found. A nominal subject is a child of its corresponding verb, annotated with the \texttt{nsubj} dependency and therefore is easily determined~(1). If the verb is a participle (\eg gerund, past participle) and its parent is an entity or pronoun, then the parent token is returned as subject~(2). When with both previous cases no subject is found, then a subroutine
(referred to as \texttt{find\_subject} 
in Algorithm~\ref{alg:find_subject}) is used with the verb's parent token as input to recursively search for a subject (3). For a token, the subroutine checks if the token itself or any child of it is a subject. Here, a token is considered to be a subject when it has the \texttt{nsubj} or \texttt{nsubjpass} dependency and when it is an entity or pronoun. If no subject is found, the subroutine is called with the parent of the token as input. This procedure terminates and returns no subject if the root of the dependency tree is reached and no subject was found. Although passive verbs are ignored for relations, as mentioned, the subroutine searches for subjects also using the \texttt{nsubjpass} dependency. This is because a subject of a passive verb can act as a subject for a non-passive verb later in the sentence (\eg \textit{"A child is seen riding a skateboard."}).

\begin{algorithm}[ht]
\small
\caption{\textit{Identify a subject for an input verb.}}
\label{alg:find_subject}
\LinesNumbered
\DontPrintSemicolon
\SetKwInput{Input}{Input}
\SetKwInput{Output}{Output}
\SetKwInput{Constants}{Constants}
\SetKwProg{Function}{Function}{:}{}

\SetKwFunction{findsubject}{find\_subject}
\SetKwFunction{findsubjectofparent}{find\_subject}
\SetKwFunction{dep}{dep}
\SetKwFunction{tag}{tag}
\SetKwFunction{childtokens}{child\_tokens}
\SetKwFunction{append}{append}
\SetKwFunction{parenttoken}{parent\_token}

\Input{Verb $v$, sentence entities $E$}
\Output{Subject of $v$}
\Function{\findsubject{$v$, $E$}}{
\textcolor{blue}{\tcp*[h]{(1) Nominal subjects}}\\
\For{$c \in$ \childtokens{$v$}}{
    \lIf{\upshape
        \dep{$c$} $=$ \texttt{nsubj} \textbf{\upshape and} ($c \in E$ \textbf{\upshape or} \tag{$c$} $=$ \texttt{PRON})}{\Return{$c$}
    }
}
\textcolor{blue}{\tcp*[h]{(2) Verb is a participle}}\\
$p =$ \parenttoken{$v$}\\
\If{\upshape
    \dep{$v$} $=$ \texttt{acl} 
    \textbf{\upshape and} 
    ($p \in E$ \textbf{\upshape or} \tag{$p$} $=$ \texttt{PRON})
    }{\Return{$p$}
}
\textcolor{blue}{\tcp*[h]{(3) Recursively search for a subject using the parent token}}\\
\Return{\findsubjectofparent{$p$, $E$}}
}
\end{algorithm}

\textbf{Search for Objects and Modifier in the Candidate Tuples}
For each verb in a candidate tuple from the second step, Algorithm~\ref{alg:find_objects} is used to determine the verb's corresponding objects and modifiers. As for subjects, at this stage an object can be either an entity or pronoun. Objects are found in two different ways. A direct object is a child of its corresponding verb, annotated with the \texttt{dobj} dependency and therefore is easily determined (1). Objects of prepositions, which are annotated with the \texttt{pobj} dependency, are not as easily determined (2). This is because objects of prepositions are not direct children of the corresponding verbs in the dependency tree. In the dependency tree, on the path between an object of preposition and its verb, there is usually one, and sometimes multiple tokens. Therefore, a recursive subroutine
(referred to as \texttt{find\_objects\_of\_prepositions()} in Algorithm~\ref{alg:find_objects}) is used to find objects of prepositions for a verb. Starting with the verb, the subroutine searches recursively for objects of prepositions by looking at child tokens. If an object of preposition is found, \ie a token which is an entity or pronoun annotated with the \texttt{pobj} dependency, then the object is returned. In addition to the object, all tokens on the path between the verb and the object are returned. These tokens are used as modifiers and therefore will be included in the corresponding element of the relation. Modifiers between the verb and an object of preposition in the dependency tree cannot be of arbitrary nature. For a relation, we limit those to tokens that have a dependency and tag combination of [\texttt{prep}, \texttt{ADP}], [\texttt{prt}, \texttt{ADP}], [\texttt{advmod}, \texttt{ADV}], [\texttt{conj}, \texttt{ADV}], or [\texttt{conj}, \texttt{ADP}]. If a coordinating conjunction, \ie a modifier with the \texttt{conj} dependency, is on the path between the verb and the object, then the relation is split up into two relations: one relation without the coordinating conjunction in the modifiers list, and one without its parent token. In addition, each modifier tagged as \texttt{ADV} needs to be included in WordNet as adverb. Since WordNet does not list prepositions, each modifier tagged as \texttt{ADP} is validated with a custom list of prepositions. Finally, the verb's leaf modifiers are determined, \ie modifiers of the verb that do not have any children (3). These are limited to tokens that have a dependency and tag combination of [\texttt{prt}, \texttt{ADP}] or [\texttt{prep}, \texttt{ADP}].

\begin{algorithm}[ht]
\small
\caption{\textit{Identify corresponding objects and modifiers for an input verb.}}
\label{alg:find_objects}
\LinesNumbered
\DontPrintSemicolon
\SetKwInput{Input}{Input}
\SetKwInput{Output}{Output}
\SetKwInput{Constants}{Constants}
\SetKwProg{Function}{Function}{:}{}

\SetKwFunction{findobjects}{find\_objects}
\SetKwFunction{findpobjs}{find\_objects\_of\_prepositions}
\SetKwFunction{dep}{dep}
\SetKwFunction{tag}{tag}
\SetKwFunction{childtokens}{child\_tokens}
\SetKwFunction{append}{append}
\SetKwFunction{parenttoken}{parent\_token}

\Input{Verb $v$, sentence entities $E$}
\Output{Tuple-list $T = [t_1,\dots]$, where a tuple $t_i = (o_i, m_i)$ contains an object $o_i$ and modifiers $m_i =[m_{i,1},\dots]$ for verb $v$}
\Function{\findobjects{$v$, $E$}}{
$T = [\ ]$\\
\textcolor{blue}{\tcp*[h]{1) Verb has direct object(s)}}\\
\For{$c$ \textbf{\upshape in} \childtokens{$v$}}{
    \lIf{\upshape
        \dep{$c$} $=$ \texttt{dobj} 
        \textbf{\upshape and} 
        ($c \in E$ \textbf{\upshape or} \tag{$c$} $=$ \texttt{PRON})
        }{$T.$\append{$(c, [\ ])$}}
}
\textcolor{blue}{\tcp*[h]{2) Verb has object(s) of preposition(s)}}\\
\For{$t$ \textbf{\upshape in} \findpobjs{$v, E$}}{
    {$T.$\append{$t$}}
}
\textcolor{blue}{\tcp*[h]{3) Search for leaf verb modifiers}}\\
\For{$c$ \textbf{\upshape in} \childtokens{$v$}}{
    \If{\upshape
        (\dep{$c$}, \tag{$c$}) $\in$ [(\texttt{prt}, \texttt{ADP}), (\texttt{prep}, \texttt{ADP})] \\ \textbf{\upshape and} \childtokens{$c$} $= \emptyset$
        }{
        \For{$t=(o, m)$ \textbf{\upshape in} $T$}{$m.$\append{$c$}}
        }
}
\Return{$T$}
}
\end{algorithm}

This almost completes the relation extraction method. So far, the subjects and objects lists of a relation each contain only a single entity or pronoun, although multiple subjects or objects may exist for it. For example, in addition to an already determined subject (\ie the head of a noun phrase), another entity can be an actor of a relation. In this case, such subject is found by looking at the child tokens of the already determined subject. If a child token is either an entity or pronoun and is annotated with the \texttt{conj} dependency, then it is also considered a subject. 

The subjects and objects lists of a definitive relation should only contain entities. However, pronouns have also been taken into account so far. As with the entity extraction method, each pronoun is checked whether it refers to an entity or not by using coreference resolution. When this is the case, then the pronoun is replaced with the corresponding entity. If for a pronoun no related entity is found, then it is simply removed from the relation's corresponding subject or object list. If a relation does not have at least one subject or object anymore, then the relation is ignored and not returned as one of the videos relations.
}

The video-level relation extraction is finished here. Event-level relations are built by assigning temporal segments to the extracted relations. 
Here, for a determined relation, the event is the temporal segment of the corresponding sentence from which the relation was extracted.

\subsection{Text Classification}
\label{sec:method_text_classification}
 
Although our proposed framework performs video classification, it predicts the category of a video only with the textual information its DVC model provides. \extended{Therefore, the framework's method for video category prediction is a text classification model.}
The motivation is to see how far a text classifier on generated video captions can correctly classify a video.
As text classifier, we adopt BERT~\cite{BERT}\extended{~(see Section~\ref{sec:text_classification_related_work}) and apply it to the captioned events, \ie the text generated by the DVC model (see Section~\ref{sec:method_event_processing})}. 
\extended{Before forwarding the text to the BERT model, its corresponding preprocessing model transforms the text to numeric token identifiers. 
The BERT model then converts the preprocessed text into embedding vectors. 
Those are fed into the classifier, a simple neural network consisting of a dropout layer and a dense layer with as many nodes as there are categories.}

\section{Experimental Apparatus}
\label{sec:experiments}

\extended{We run experiments for videos that are included in the ActivityNet Captions~\cite{DVCPioneerANetCaptions} dataset. We evaluate each of our semantic metadata extraction methods separately by collecting multiple metrics and compare results for our framework in two settings, \ie while using two different DVC models. With MT\footnote{\label{footnote:github_MT}MT is available at: \url{https://github.com/salesforce/densecap}.} (End-to-End Dense Video Captioning with Masked Transformer)~\cite{MT} and PDVC\footnote{\label{footnote:github_PDVC}PDVC is available at: \url{https://github.com/ttengwang/PDVC}.} (End-to-End Dense Video Captioning with Parallel Decoding)~\cite{PDVC}, we use two different DVC models that provide high-quality captioning results.}

\subsection{Datasets}
\label{sec:experiments/datasets}

We introduce the datasets to evaluate the DVC models and the different tasks of our metadata extraction framework.

\subsubsection*{Dense Video Captioning}
We use the large-scale benchmark dataset ActivityNet Captions\extended{\footnote{ActivityNet Captions is available at: \url{https://cs.stanford.edu/people/ranjaykrishna/densevid/}.}}~\cite{DVCPioneerANetCaptions} to train and evaluate the dense video captioning models. 
It consists of 20k YouTube videos of various human activities split up into train/val/test sets of 0.5/0.25/0.25\extended{, with the test set currently being withheld}. 
Each video is annotated with captioned events, each consisting of a descriptive sentence and a specific temporal segment to which the description refers. 
On average, each video is annotated with 3.65 temporally-localized sentences. 
Each captioned event on average covers 36 seconds and is composed of 13.5 words. Temporal segments in the same video can overlap in time, which enables DVC models to learn complex events and relations. 
\extended{
  \footnote{Note that ActivityNet Captions consists of two different annotation files for the validation
set,  \texttt{val\_1.json} and \texttt{val\_2.json}.
Most of the times, we use the first annotation file. 
Throughout this paper, we always make explicitly clear which annotation files of the ActivityNet Captions dataset validation set we are using.}
}

\subsubsection*{Entity, Property, and Relation Extraction}

The information that ActivityNet Captions provides for each video is limited to captioned events, \ie pairs of temporal segments and sentences. To be able to evaluate the entity, property, and relation extraction methods, we need information about depicted entities, properties of entities, and relations of videos. 
For this purpose, we utilize the gold standard captioned events of ActivityNet Captions' validation videos\extended{~(using only the first of the two annotation files)}.
We extract entities, their properties, and relations from the captioned events, and treat the results as gold standard for semantic metadata extraction. 
We generate five different datasets for videos in the ActivityNet Captions validation set: 
each one for event-level entities, video-level entities, entity-property pairs, event-level relations, and video-level relations. 
Using these datasets, we evaluate the framework's ability to extract the semantic metadata and compare it to metadata extracted from the captioned events generated by the DVC models.

\subsubsection*{Text Classification}
We build a new dataset to train and evaluate our text classification method. Here, we take advantage of the fact that ActivityNet Captions consists of YouTube videos. For each video of the ActivityNet Captions' train and validation sets, we query the corresponding category from the YouTube Data API. 
We were able to query the category of $12,579$ ActivityNet Captions videos (on March 20, 2022). 
\extended{The remaining videos are no longer (publicly) available on YouTube.}
We split the videos in train/val/test of 0.6/0.2/0.2, while ensuring that the category distribution is the same for all splits. 
\extended{An overview of the YouTube video categories and their frequencies in our dataset is shown in Table~\ref{tab:categories_dataset}.

\begin{table}
\small
    \centering
    \begin{tabular}{l|l|r|r}
        \toprule
        ID & Category & \# videos & \% \\
        \midrule
        1 & Film \& Animation        & 213  & 1.69\\
        2 & Cars \& Vehicles         & 191  & 1.52\\
        10 & Music                   & 637  & 5.06\\
        15 & Pets \& Animals         & 342  & 2.72\\
        17 & Sport                   & 3,920 & 31.16\\
        19 & Travel \& Events        & 526  & 4.18\\
        20 & Gaming                  & 78   & 0.62\\
        22 & People \& Blogs         & 2,098 & 16.68\\
        23 & Comedy                  & 645  & 5.13\\
        24 & Entertainment           & 1,414 & 11.24\\
        25 & News \& Politics        & 226  & 1.80\\
        26 & How-to \& Style         & 1,661 & 13.20\\
        27 & Education               & 414  & 3.29\\
        28 & Science \& Technology   & 146  & 1.16\\
        29 & Non-profits \& Activism & 68   & 0.54\\
        \midrule 
        \multicolumn{1}{c}{} & \multicolumn{1}{c}{} & \multicolumn{1}{r}{12,579} & \multicolumn{1}{r}{100\%} \\
        \bottomrule
    \end{tabular}
    \caption[Categories of YouTube videos and their frequency in our classification dataset.]{Categories of YouTube videos and their frequency in our classification dataset.}
    \label{tab:categories_dataset}
\end{table}
}

\subsection{Procedure}
\label{sec:experiments/procedure}

\subsubsection*{Dense Video Captioning} 
We train MT and PDVC using the ActivityNet Captions dataset. For action recognition, both models adopt the same action recognition network, a pre-trained temporal segment network~\cite{TSN}, to extract frame-level features. To ensure consistency, we evaluate both models on the ActivityNet Captions validation set (using both annotation files\extended{~of the validation set}) and compare the performances with those reported in the corresponding works. 

For our proposed semantic metadata extraction methods, the number of captioned events forwarded to each method is an important parameter. With increasing number of captioned events forwarded to an extraction method, it can extract more semantic information. In this work, we denote the number of captioned events that are generated by a DVC model and forwarded to a specific method as $|E|$. MT and PDVC, the dense video captioning models of our choices, internally calculate confidence scores for their generated captioned events. If the number of considered captioned events is limited, then the generated captioned events with the highest confident scores are used. For high values of $|E|$, DVC models tend to produce identical sentences for different temporal segments. However, our property extraction, and video-level entity and relation extraction methods rely mainly on the textual information that the captioned events provide. Therefore, captioned events with duplicate sentences do not provide further semantic information. Because of that, we only forward events with distinct captions to these methods. We denote the number of distinct captioned events that are forwarded to a specific method as $|dist(E)|$. If two events share the exact same caption, then the event with the higher confidence score is forwarded. We finally evaluate MT and PDVC with $|E|$ set to 10, 25, 50, and 100, and $|dist(E)|$ set to 1, 3, 10, and 25.

\subsubsection*{Entity, Property, and Relation Extraction}
For both trained dense video captioning models, MT and PDVC, we extract event-level entities and relations from captioned events that they generate for ActivityNet Captions validation videos with $|E|$ set to 10, 25, 50, and 100. Video-level entities and relations, and entity-property pairs are extracted with $|dist(E)|$ set to 1, 3, 10, and 25. We then evaluate the entity, property, and relation extraction methods by comparing the extracted semantic metadata with our generated gold standards for entities, entity-property pairs, and relations.

For the evaluation of video-level entity extraction, we introduce the entity frequency threshold $f$. 
We extract video-level entities from ActivityNet Captions' train and validation videos. 
The frequency of a video-level entity is the number of different videos in which it occurs. 
We evaluate the generated video-level entities with $f$ set to 0, 10, 25, and 50, meaning that the gold standard only contains those entities which have a frequency higher than the frequency threshold. 
Thus, a larger $f$ means that the DVC models are given a higher chance to learn and reproduce the entities in the videos.

\extended{
\begin{table}
   \small
    \centering
        \begin{tabular}{l|cccc}
            \toprule
            \multicolumn{1}{c|}{Frequency threshold $f$} & 0 & 10 & 25 & 50 \\
            \midrule
            \# Entities & 5,428 & 1,358 & 794 & 468\\
            \bottomrule
        \end{tabular}
    \caption[Number of different video-level entities in the ActivityNet Captions dataset per entity frequency threshold.]{Different video-level entities in the ActivityNet Captions dataset per entity frequency threshold. The entity frequency is the number of different ActivityNet Captions videos in which it occurs.}
    \label{tab:statistics_entity_frequency_threshold}
\end{table}
}

\subsubsection*{Text Classification}
We train and evaluate the text classification model of our framework in three different settings: using the captioned events generated by the MT and PDVC model, respectively, and using the gold standard captioned events provided by ActivityNet Captions\extended{ (using only the first of the two annotation files for validation videos)}. 
This allows us to analyze how useful the automatically generated captioned events are for text classification compared to gold standard captioned events. For each data input, we set $|dist(E)|$ to 10. 
\extended{Note that the text classification model is also trained and evaluated on videos for which the DVC models were already trained on. This is because our dataset for classification contains train and validation videos of ActivityNet Captions, which are randomly split up into train/val/test sets. 
The reason we also include ActivityNet Captions' train videos into the text classification dataset is to ensure that there are enough videos in all categories to train, validate, and test the text classification method on.}

\subsection{Hyperparameter Optimization}
\label{sec:experiments/hyperparameters}

\extended{\subsubsection*{Dense Video Captioning}}
For the training of the DVC models, we use largely the same parameters that were used to train the models in their original works (refer to MT\extended{\footref{footnote:github_MT}}~\cite{MT} and PDVC\extended{\footref{footnote:github_PDVC}}~\cite{PDVC} or their latest codebases). For MT, we have to switch to a batch size of 84 and a learning rate of 0.06 to make sure that the model training converges. In the default configuration, PDVC generates a maximum of 10 events per video. We change the corresponding parameter such that 100 captioned events are generated to ensure that there is enough information for our metadata extraction methods to work on. For both, MT and PDVC, we use the models' best states w.r.t. their METEOR score performances (see Section~\ref{sec:experiments/measures}). For MT, this was achieved after 34 epochs, and for PDVC after 13 epochs.

\extended{\subsubsection*{Entity, Property, and Relation Extraction and Text Classification}}
For the entity, property, and relation extraction methods, no hyperparameter optimization is necessary.
In our text classification model, we use an uncased BERT model\extended{ consisting of 12 transformer blocks, a hidden size of 768, and 12 attention heads\footnote{From: \url{https://tfhub.dev/tensorflow/bert_en_uncased_L-12_H-768_A-12/4}.}}. 
Each input text is truncated to 128 tokens. For training, we use cross-entropy loss and the Adam optimizer with 10\% warmup steps. We use class weights to help the model learn on the imbalanced data of our classification dataset\extended{~(see Table \ref{tab:categories_dataset})}. While using captioned events from ActivityNet Captions as training data, we perform grid search over dropout rates in $\{0.0, 0.1, \dots, 0.5\}$, maximum number of training epochs in $\{1, 2, \dots, 5\}$, learning rates between $5e$-$6$ and $1e$-$4$, and a batch sizes in $\{1, 2, 4, 8\}$. The lowest validation loss was achieved after epoch 2 while using dropout of 0.1, 2 maximum training epochs, learning rate of $2e$-$5$, and batch size of 4. Using this parameter configuration, we train the model with each data input separately. We repeat the training three times \extended{for each data input~}and select the model that achieved the lowest final validation loss.

\subsection{Measures and Metrics}
\label{sec:experiments/measures}

\subsubsection*{Dense Video Captioning}
To evaluate the dense video captioning models MT and PDVC, we use the official evaluation toolkit provided by ActivityNet Captions Challenge 2018\footnote{See: \url{https://github.com/ranjaykrishna/densevid_eval/}}
that measures the capability to localize and describe events in videos. 
For evaluation of dense video captioning, we report METEOR~\cite{METEOR} and BLEU~\cite{BLEU} scores. 
\extended{BLEU is a precision-based metric, which accounts for precise matching of n-grams between a generated and ground truth sentence. METEOR, however, first creates an alignment by comparing exact tokens, stemmed tokens, and paraphrases of the prediction and ground truth sentence. It also takes semantically similar matches into account by using WordNet for synonym analysis~\cite{SurveyVideoCaptioning1}.}
Using temporal Intersection over Union (tIoU) thresholds at \{0.3, 0.5, 0.7, 0.9\} for captioned events, we report recall and precision of their temporal segments and METEOR and BLEU scores of their sentences. Given a tIoU threshold, if the event proposal has a segment overlapping larger than the threshold with any gold standard segment, the metric score is computed for the generated sentence and the corresponding gold standard sentence. Otherwise, the metric score is set to 0. The scores are then averaged across all event proposals and finally averaged across all tIoU thresholds.

\subsubsection*{Entity, Property, and Relation Extraction}
For property extraction and video-level entity and relation extraction, we measure micro-averaged precision, recall, and their harmonic mean, the F1 score. Similarly to the evaluation of DVC models, when evaluating entity and relation extraction on event-level, we also take the quality of the predictions' temporal segments into account. Here, we compute micro-averaged precision and recall across all videos using tIoU thresholds at \{0.3, 0.5, 0.7, 0.9\}, and then average the results across all thresholds. Additionally, we report the F1 score of averaged precision and recall values.

For the evaluation of entities, entity-property pairs, and relations, we use WordNet to compare words in terms of whether they are synonyms of each other or not. This is fair as the variety of extracted semantic metadata is large. For example, \textit{"stand up"} is considered a correct prediction for the verb \textit{"get up"}. 
However, this means that for a video the number of predictions that are treated as correct (the set $TP_p$) and the number of gold standard targets (denoted as $TP_g$) is not necessarily the same.
With other words, the gold standard is enriched with synonyms.
For example, $|TP_p |=2$ and $|TP_g |=1$ when accepting \textit{person} and \textit{individual} as synonymously correct predictions for $TP_g = \{$\textit{person}$ \}$.
To ensure the validity of precision (proportion of correct predictions) and recall (proportion of correctly predicted targets), we use $TP_p$ for calculating precision and $TP_g$ for recall, respectively.

All entities, properties, and relations are compared using their word lemmas. 
On video-level, this means that potential duplicates of entities and relations are removed, \ie they are unique\extended{ (after lemmatization)}. 
For example, $\{$\textit{man, men}$\}$ results in $\{$\textit{man}$\}$. 

\extended{The following rules define when a prediction for an entity, entity-property pair, or relation is correct:

\textbf{Entities.} A video-level entity $e_s$ predicts $e_g$ correctly when for its noun holds $n_s \in SYN(n_g)$. Similarly to the evaluation of DVC models, in the evaluation of event-level entity extraction, we also take the entities' temporal segments into account. Given a tIoU threshold, an event-level entity $e_s$ predicts $e_g$ correctly, if 1) $n_s \in SYN(n_g)$ and 2) $e_s$ has a segment overlapping with the segment of $e_g$ larger than the threshold.

\textbf{Properties.} An entity-property pair $(e_p, a_p)$ predicts $(e_g, a_g)$ correctly, if $e_p \in SYN(e_g)$ and $a_p \in SYN(a_g)$.

\textbf{Relations.} On video-level, a relation $r_p$ predicts a target relation $r_g$ correctly, if 1) \textit{subjects}$_p$ contains at least one entity that predicts any entity of \textit{subjects}$_g$, and 2) \textit{objects}$_p$ contains at least one entity that predicts any entity of \textit{objects}$_g$. In addition to that, 3) the verb \textit{verb}$_p$, or any combination of it with one of its modifiers in \textit{modifier}$_p$ has to predict the \textit{verb}$_g$, or any combinations of it with one of its modifiers in \textit{modifier}$_g$. Again, if a synonym of a target subject, verb or object is predicted, then the corresponding prediction is correct. If we use the above rules, then, \eg \textit{([man, woman], go, [in], building)} is a correct prediction of \textit{(man, enter, [], building)}. For the evaluation of event-level relation extraction, we take the relations' temporal segments into account in the same way as we do for entities. 
}

\subsubsection*{Text Classification}
\extended{Precision, recall, and F1 score are vital metrics for the evaluation of (single-label, multi-class) classification methods on unbalanced test sets~\cite{TextClassificationSurvey1}.}
We report weighted\extended{ (with weights equal to the category probability)} and macro-averages of precision, recall, and F1 score across all categories. 
\extended{This includes the calculation of accuracy, as this is equal to the weighted recall.}

\extended{
\subsection{Implementation}
We implemented our framework in Python, using spaCy\footref{footnote:spaCy} as the language parser of our choice and TensorFlow\footnote{TensorFlow is available at: \url{https://github.com/tensorflow/tensorflow}.} for the implementation of the text classification method. Within spaCy, we are bound to use version 2.3.7 and the \texttt{en\_core\_web\_lg} \texttt{-2.3.1} language model\footnote{The spaCy model that we use is available at: \url{https://github.com/explosion/spacy-models/releases/tag/en_core_web_lg-2.3.1}.}. This is because NeuralCoref\footref{footnote:NeuralCoref}, a pipeline extension for spaCy which we use for coreference resolution, is not compatible with versions 3.0.0+ yet. The language model achieves accuracies for fine grained POS tags of 97.22 and labeled dependencies of 90.28. As stated before, WordNet~\cite{WordNet} is used at various points for lemmatizing words and to ensure that extracted semantic metadata consists of linguistically correct English nouns, verbs, adjectives, and adverbs\footnote{When using the WordNet lemmatizer, we make exceptions for few nouns and verbs which WordNet does not lemmatize as expected. Example: WordNet lemmatizes \textit{men} to \textit{men}. We make a hard-coded exception and lemmatize \textit{men} to \textit{man}.}. In addition to that, WordNet is used for the evaluation of semantic metadata extraction methods to compare words in terms of whether they are synonyms of each other or not. 
}

\section{Results}
\label{sec:results}

\extended{Given the experimental setup from Section~\ref{sec:experiments}, we evaluate the
results of dense video captioning, entity, property, and relation extraction, and text classification. 
For further experimental results refer to Section~\ref{sec:extended_results} of the Supplementary Materials. For example, Table~\ref{tab:statistics_frequent_semantic_metadata} lists the video-level entities, entity-property pairs, and video-level relations that are most frequently extracted by our framework for ActivityNet Captions videos.}

\begin{table*}[ht]
\centering
    \begin{subtable}[h]{\textwidth}
 \scriptsize
    \centering
        \begin{tabular}{l|N{33pt}||N{27pt}N{33pt}|N{21pt}N{21pt}N{21pt}}
            \toprule
            \multicolumn{7}{c}{\textit{Event Localization \& Dense Video Captioning}} \\
            \toprule
            DVC & & avg. & avg. & \multicolumn{3}{c}{2018 eval. toolkit} \\
            Model & $|E|$ & Recall &  Precision & B@3 & B@4 & M \\
            \midrule
            MT & 10 & 43.61 & 48.96 & 2.18 & 1.02 & 5.89 \\ 
             & 25 & 56.55 & 46.15 & 2.33 & 1.14 & 5.74 \\ 
             & 50 & 67.70 & 41.60 & 2.31 & 1.13 & 5.35 \\  
             & 100 & 76.33 & 34.78 & 2.12 & 1.04 & 4.68 \\ 
            \midrule
            PDVC & 10 & 61.88 & 45.41 & 3.10 & 1.59 & 6.34 \\ 
             & 25 & 73.81 & 36.89 & 2.54 & 1.23 & 5.17 \\ 
             & 50 & 78.76 & 25.75 & 1.92 & 0.90 & 3.88 \\ 
             & 100 & 82.24 & 15.63 & 1.36 & 0.63 & 2.63 \\ 
            \bottomrule
        \end{tabular}
        \caption{Results for varying numbers of generated captioned events $|(E)|$.}
        \label{tab:results_dvc_per_E_non_distinct}
    \end{subtable}

\vspace{2mm}

    \begin{subtable}[h]{\textwidth}
     \scriptsize
    \centering
        \begin{tabular}{l|N{33pt}||N{27pt}N{33pt}|N{21pt}N{21pt}N{21pt}}
            \toprule
            \multicolumn{7}{c}{\textit{Event Localization \& Dense Video Captioning}} \\
            \toprule
            DVC & & avg. & avg. & \multicolumn{3}{c}{2018 eval. toolkit} \\
            Model & $|dist(E)|$ & Recall &  Precision & B@3 & B@4 & M \\
            \midrule
            MT & 1 & 22.04 & 50.99 & 1.55 & 0.68 & 5.37 \\ 
            & 3 & 32.72 & 49.65 & 1.97 & 0.90 & 5.76 \\ 
            & 10 & 50.08 & 44.75 & 2.22 & 1.05 & 5.58 \\ 
            & 25 & 63.67 & 36.23 & 2.05 & 0.97 & 4.80 \\ 
            \midrule
            PDVC & 1 & 20.05 & 49.63 & 2.52 & 1.30 & 5.86 \\ 
            & 3 & 40.28 & 48.26 & 2.98 & 1.54 & 6.46 \\ 
            & 10 & 62.38 & 44.10 & 2.85 & 1.41 & 6.01 \\ 
            & 25 & 71.62 & 31.69 & 2.14 & 1.00 & 4.54 \\ 
            \bottomrule
        \end{tabular}
        \caption{Results for varying numbers of generated distinct captioned events $|dist(E)|$.}
        \label{tab:results_dvc_per_E_distinct}
    \end{subtable}

\caption[Event localization and dense video captioning results for different numbers of generated captioned events.]{Event localization and dense video captioning results of MT and PDVC for different numbers of generated captioned events on the ActivityNet Captions validation set.
\extended{Only the first of the two gold standard annotation files of the validation set is used, as this contains the data from which we generate our gold standards for semantic metadata.}
We report recall and precision of temporal segments, METEOR (M), and BLUE@N (B@N)  of generated captioned events.} 
\label{tab:results_dvc_per_E}
\end{table*}

\subsection{Dense Video Captioning}
\label{sec:results/dense_video_captioning}

We introduced the parameters $|E|$ and $|dist(E)|$, which are primarily used in our framework to control the amount of information that is forwarded from the DVC model to the semantic metadata extraction methods. 
Using these parameters, we are also able to compare the DVC models in a fair way, meaning that their performances are evaluated while generating an equal number of captured events. Table~\ref{tab:results_dvc_per_E} shows the event localization and dense video captioning performances of MT and PDVC with respect to $|E|$ and $|dist(E)|$, the number of (distinct) captioned events that are generated by each DVC model and used for the evaluation. 
For event localization, PDVC constantly achieves better recall than MT, while MT constantly outperforms PDVC in terms of precision. For both models, higher values of $|E|$ and $|dist(E)|$ results in better recall and worse precision performance in event localization. Looking at the results for dense video captioning, we can state that the METEOR performances of both models degrade with increasing number of captioned events, both distinct and non-distinct, except for when increasing $|dist(E)|$ from $1$ to $3$. 
When increasing $|E|$ from $10$ to $100$, the METEOR performance of PDVC drops by $3.71$ points in total, while the METEOR performance of MT drops by $1.21$ points. 
This is a consequence of the declining precision for event localization for higher $|E|$, which affects the PDVC model more heavily than the MT model.

\subsection{Entity Extraction}
Table~\ref{tab:results_video_level_entities} shows precision, recall, and F1 score performances of our framework for \textbf{video-level entity} extraction. 
\extended{As for all semantic metadata types, we evaluate our framework in two different settings: while using the MT model, and while using the PDVC model for the generation of semantic information. }
We evaluate the extracted video-level entities with entity frequency threshold $f$ set to 0, 10, 25, and 50\extended{, meaning that the gold standard only contains those video-level entities that have a frequency higher than the threshold}. 
One observation is that, with increasing $|dist(E)|$, precision decreases and recall improves for both models and all thresholds. For both DVC models and all thresholds, the best F1 score performance is achieved with $|dist(E)| = 10$. This indicates a limited level of semantic information that any further generated captioned events provide. 
For higher entity frequency thresholds, for both DVC models and all $|dist(E)|$, our framework is able to predict video-level entities with improved recall, at the cost of only slightly lower precision. When our framework is using the PDVC model, it achieves better precision and F1 scores for all $|dist(E)|$. 
The framework achieves better recall performances when using the MT model for $|dist(E)|$ set to 10 and 25. Overall, for video-level entity extraction, our framework achieves its highest F1 scores when using the PDVC model with $|dist(E)|$ set to 10. Here, the achieved F1 scores range from 31.27 for $f=0$ to 34.21 for $f=50$.

\begin{table*}[ht]
\scriptsize
\centering   
    \begin{tabular}{l|N{31pt}||N{21pt}N{21pt}N{21pt}N{21pt}|N{21pt}N{21pt}N{21pt}N{21pt}|N{21pt}N{21pt}N{21pt}N{21pt}}
        \toprule
        \multicolumn{14}{c}{\textit{\textbf{Video-level Entity} Extraction}} \\
        \toprule
        DVC &  & \multicolumn{4}{c|}{Precision(@$f$)} & \multicolumn{4}{c|}{Recall(@$f$)} & \multicolumn{4}{c}{F1(@$f$)} \\
        Model & $|dist(E)|$ & 0 & 10 & 25 & 50 & 0 & 10 & 25 & 50 & 0 & 10 & 25 & 50\\
        \midrule
        MT & 1 & 39.94 & 39.88 & 39.66 & 38.91 & 15.38 & 16.49 & 17.57 & 18.99 & 22.21 & 23.33 & 24.35 & 25.52 \\
         & 3 & 33.91 & 33.83 & 33.61 & 32.78 & 23.44 & 25.12 & 26.72 & 28.70 & 27.72 & 28.83 & 29.77 & 30.60 \\
         & 10 & 26.44 & 26.37 & 26.16 & 25.43 & 32.59 & 34.90 & 37.06 & 39.67 & 29.20 & 30.04 & 30.67 & 30.99 \\
         & 25 & 20.76 & 20.69 & 20.49 & 19.82 & 40.69 & 43.55 & 46.16 & 49.15 & 27.49 & 28.05 & 28.38 & 28.25 \\
        \midrule
        PDVC & 1 & 45.13 & 45.05 & 44.91 & 44.43 & 15.44 & 16.56 & 17.68 & 19.28 & 23.01 & 24.22 & 25.37 & 26.89 \\
         & 3 & 39.01 & 38.95 & 38.83 & 38.28 & 23.33 & 25.03 & 26.72 & 29.04 & 29.20 & 30.48 & 31.66 & 33.03 \\
         & 10 & 31.77 & 31.71 & 31.60 & 31.03 & 30.78 & 33.01 & 35.21 & 38.11 & 31.27 & 32.35 & 33.30 & 34.21 \\
         & 25 & 26.30 & 26.25 & 26.14 & 25.61 & 36.18 & 38.79 & 41.33 & 44.66 & 30.46 & 31.31 & 32.02 & 32.55 \\
        \bottomrule
    \end{tabular}
   \caption[Experimental results for video-level entity extraction.]{Video-level entity extraction using dense video captioning models MT and PDVC. 
   We report precision, recall, and F1 score \extended{on our gold standard for video-level entities }for different numbers of generated distinct captioned events $|dist(E)|$ and entity frequency thresholds $f$.\\} 
    \label{tab:results_video_level_entities}
\end{table*}

\begin{table*}[ht]
    \scriptsize
\centering
    \begin{tabular}{l|N{25pt}||N{22pt}N{22pt}N{22pt}N{22pt}N{22pt}|N{22pt}N{22pt}N{22pt}N{22pt}N{22pt}|N{22pt}}
        \toprule
        \multicolumn{13}{c}{\textit{\textbf{Event-level Entity} Extraction}} \\
        \toprule
         &  & \multicolumn{5}{c|}{Precision(@tIoU)} & \multicolumn{5}{c|}{Recall(@tIoU)} & \\
        DVC Model & $|E|$ & 0.3 & 0.5 & 0.7 & 0.9 & Avg & 0.3 & 0.5 & 0.7 & 0.9 & Avg & F1\\
        \midrule
        MT & 10 & 28.97 & 20.08 & 8.55 & 1.26 & 14.72 & 21.11 & 15.31 & 9.59 & 2.94 & 12.24 & 13.37 \\
         & 25 & 28.12 & 18.44 & 7.40 & 1.03 & 13.75 & 26.48 & 20.66 & 14.29 & 5.58 & 16.75 & 15.10 \\
         & 50 & 26.30 & 15.93 & 6.10 & 0.83 & 12.29 & 31.15 & 25.32 & 18.69 & 8.47 & 20.91 & 15.48 \\
         & 100 & 23.31 & 12.70 & 4.52 & 0.59 & 10.28 & 36.35 & 30.05 & 22.89 & 11.04 & 25.08 & 14.58 \\
        \midrule
        PDVC & 10 & 30.47 & 19.79 & 9.11 & 2.53 & 15.47 & 26.36 & 21.13 & 14.52 & 5.87 & 16.97 & 16.19 \\
         & 25 & 27.16 & 15.83 & 6.41 & 1.53 & 12.73 & 31.14 & 26.34 & 20.01 & 8.21 & 21.42 & 15.97 \\
         & 50 & 20.75 & 10.49 & 3.84 & 0.86 & 8.98 & 33.66 & 28.89 & 22.40 & 8.75 & 23.42 & 12.98 \\
         & 100 & 13.76 & 6.36 & 2.22 & 0.48 & 5.71 & 35.35 & 30.65 & 24.00 & 9.02 & 24.75 & 9.28 \\
        \bottomrule
    \end{tabular}
    \caption[Experimental results for event-level entity extraction.]{Event-level entity extraction results. 
    Reported are precision and recall \extended{on our gold standard for event-level entities }for different numbers of generated distinct captioned events 
    $|E|$    
    and tIoU thresholds. 
    The F1 scores are \extended{calculated using }the averages of precision and recall across all thresholds.}
    \label{tab:results_event_level_entities}
\end{table*}

Table~\ref{tab:results_event_level_entities} shows the results of \textbf{event-level entity} extraction of our framework. Depending on the framework's used DVC model and the number of generated captioned events $|E|$, we report precision and recall for different temporal Intersection over Union (tIoU) thresholds. F1 score is calculated using the averages of precision and recall. Note that for a prediction to be correct for an event-level entity, a condition is that its temporal segment overlaps with the gold standard entity's temporal segment larger than the tIoU threshold. Therefore, precision and recall decreases at higher tIoU thresholds. Regardless of the DVC model used, the framework's precision decreases for higher $|E|$, while at the same time it benefits in recall performance. When using the PDVC model, the framework's precision drops more (1.13 on average) when increasing $|E|$ from 10 to 100 as compared to when it is using the MT model (0.57 on average). Note that we made the same observation when we evaluated the event localization and dense video captioning performances of the DVC models. For event localization, the PDVC model could not convert the large drops in precision for higher $|E|$ into better recall performance. Thus, on the one hand, this results in worse dense video captioning performance. On the other hand, when our framework is using the PDVC model for event-level entity extraction, this results in degrading F1 score performance for higher $|E|$. Still, our framework achieves its highest F1 score with 16.19 when it uses the PDVC model with $|E| = 10$. On the other hand, when using the MT model, the highest achieved F1 score is 15.48 for $|E| = 50$.

\subsection{Property Extraction}
The results of the extraction of entity-property pairs with our framework are shown in Table~\ref{tab:results_properties}. In general, increasing $|dist(E)|$ leads to drops in precision, however, our framework benefits from much improved recall. Consequently, our framework achieves its highest F1 scores for $|dist(E)| = 25$. In contrast to video-level entity extraction where highest F1 scores were achieved for $|dist(E)| = 10$, we can observe that increasing $|dist(E)|$ from 10 to 25 leads to even better property extraction performance with respect to the F1 score, indicating that the DVC models still provide meaningful semantic information about the properties of entities when many captioned events are generated. For all $|dist(E)|$, our framework achieves better recall and F1 scores when using the MT model for captioned events generation and better precision when using the PDVC model. The highest achieved precision is 9.08 when using PDVC with $|dist(E)| = 1$. Using the MT model with $|dist(E)| = 25$ results in the framework's highest achieved recall (8.86) and F1 score (4.94).

\begin{table}
\scriptsize
\centering
    \begin{tabular}{l|N{35pt}||N{19pt}N{19pt}N{19pt}}
        \toprule
        \multicolumn{5}{c}{\textit{\textbf{Property} Extraction}} \\
        \toprule
        DVC Model & $|dist(E)|$ & Prec. & Rec. & F1 \\
        \midrule
        MT & 1 & 6.48 & 0.82 & 1.45 \\
         & 3 & 6.64 & 1.85 & 2.90 \\
         & 10 & 5.66 & 3.54 & 4.36 \\
         & 25 & 4.53 & 5.43 & 4.94 \\
        \midrule
        PDVC & 1 & 9.08 & 0.72 & 1.34 \\
         & 3 & 8.76 & 1.61 & 2.72 \\
         & 10 & 6.96 & 2.68 & 3.87 \\
         & 25 & 4.79 & 3.76 & 4.21 \\
        \bottomrule
    \end{tabular}
    \caption{Results for the extraction of entity-property pairs.}
    \label{tab:results_properties}
\end{table}

\subsection{Relation Extraction}

Table~\ref{tab:results_video_level_relations} shows the results of video-level relation extraction with our framework. 
\extended{Again, an increase in $|dist(E)|$ leads to drop in precision, while recall performance improves.}
In general, using the PDVC model leads to better precision performance except for $|dist(E)| = 1$, while using the MT model leads to better recall performance except for $|dist(E)| = 3$. For both DVC models, our framework achieves its highest F1 scores for $|dist(E)| = 10$. 
This suggests that any larger number of generated captioned events can provide more semantic information for relations only at a higher cost of precision, the same observation as we made for video-level entity extraction. However, for property extraction, the highest F1 scores were achieved for $|dist(E)| = 25$. The highest achieved F1 score of our framework for video-level relation extraction is 5.02 while using the PDVC model with $|dist(E)| = 10$.

The results for event-level relation extraction are shown in Table~\ref{tab:results_event_level_relations}. Very similar observations can be made as for event-level entity extraction. The framework's precision decreases for higher $|E|$ while benefiting in recall performance. When using the PDVC model, the framework's precision performance suffers more for higher $|E|$ compared to when it is using the MT model. As observed before for event-level entity extraction, this is not converted into much higher recall, resulting in a degradation of the framework's F1 score performance. Therefore, for $|E|$ set to 50 and 100, \ie high numbers of generated captioned events, our framework achieves its highest F1 scores when using the MT model, while for $|E|$ set to 10 and 25 the highest F1 scores are achieved when using the PDVC model. 
The event-level relation extraction achieves its best performance with respect to F1 score when it uses the PDVC model with $|E|=10$.

\begin{table}
\scriptsize
\centering
    \begin{tabular}{l|N{35pt}||N{19pt}N{19pt}N{19pt}}
        \toprule
        \multicolumn{5}{c}{\textit{\textbf{Video-level Relation} Extraction}} \\
        \toprule
        DVC Model & $|dist(E)|$ & Prec. & Rec. & F1 \\
        \midrule
        MT & 1 & 5.76 & 2.07 & 3.04 \\
         & 3 & 4.88 & 4.07 & 4.44 \\
         & 10 & 3.64 & 6.61 & 4.70 \\
         & 25 & 2.89 & 8.86 & 4.35 \\
        \midrule
        PDVC & 1 & 5.64 & 2.06 & 3.02 \\
         & 3 & 5.02 & 4.18 & 4.56 \\
         & 10 & 4.09 & 6.50 & 5.02 \\
         & 25 & 3.47 & 8.36 & 4.91 \\
        \bottomrule
    \end{tabular}
    \caption{Results for video-level relation extraction.}
    \label{tab:results_video_level_relations}
\end{table}

\begin{table*}[ht]
\scriptsize
\centering
    \begin{tabular}{l|N{25pt}||N{19pt}N{19pt}N{19pt}N{19pt}N{19pt}|N{19pt}N{19pt}N{19pt}N{19pt}N{19pt}|N{19pt}}
        \toprule
        \multicolumn{13}{c}{\textit{\textbf{Event-level Relation} Extraction}} \\
        \toprule
         & & \multicolumn{5}{c|}{Precision(@tIoU)} & \multicolumn{5}{c|}{Recall(@tIoU)} & \\
        DVC Model & $|E|$ & 0.3 & 0.5 & 0.7 & 0.9 & Avg & 0.3 & 0.5 & 0.7 & 0.9 & Avg & F1\\
        \midrule
        MT & 10 & 3.76 & 2.52 & 1.08 & 0.14 & 1.87 & 3.81 & 2.81 & 1.73 & 0.42 & 2.20 & 2.02 \\
         & 25 & 3.61 & 2.28 & 0.93 & 0.12 & 1.74 & 4.89 & 3.90 & 2.56 & 0.87 & 3.05 & 2.22 \\
         & 50 & 3.35 & 1.99 & 0.77 & 0.10 & 1.55 & 5.87 & 4.78 & 3.29 & 1.32 & 3.81 & 2.20 \\
         & 100 & 2.96 & 1.59 & 0.56 & 0.07 & 1.30 & 7.06 & 5.60 & 3.91 & 1.63 & 4.55 & 2.02 \\
        \midrule
        PDVC & 10 & 3.64 & 2.33 & 1.14 & 0.32 & 1.86 & 4.87 & 3.87 & 2.63 & 1.06 & 3.11 & 2.33 \\
         & 25 & 3.33 & 1.92 & 0.80 & 0.20 & 1.56 & 6.01 & 4.95 & 3.49 & 1.37 & 3.95 & 2.24 \\
         & 50 & 2.66 & 1.33 & 0.50 & 0.11 & 1.15 & 6.53 & 5.37 & 3.81 & 1.41 & 4.28 & 1.81 \\
         & 100 & 1.75 & 0.82 & 0.30 & 0.07 & 0.73 & 6.75 & 5.53 & 3.91 & 1.41 & 4.40 & 1.25 \\
        \bottomrule
    \end{tabular}
    \caption[Experimental results for event-level relation extraction.]{Experimental results for event-level relation extraction.}
    \label{tab:results_event_level_relations}
\end{table*}

\subsection{Text Classification}
\label{sec:results/category}
Finally, Table~\ref{tab:results_classification} shows the weighted and macro-averages of precision, recall, and F1 score performances of the framework's text classification method, while trained and evaluated in three different settings: using the captioned events generated by the MT or PDVC model, respectively, and using the captioned events provided by ActivityNet Captions. The most noteworthy observation is that the classification performance of our framework, when using captioned events generated by MT and PDVC, is not far from the classification performance that is achieved when using gold standard captioned events of ActivityNet Captions. Therefore, we can state that the DVC models generate specific semantic information for videos of different categories at a similar level as the captioned events of the gold standard provide. This is important for the text classifier to categorize videos successfully. 

When using automatically generated captioned events for video classification, our framework achieves its best performances for all metrics, both weighted and macro-averaged, when using the PDVC model. Here, for weighted precision, recall, and F1 score, our framework performs around 2 points better as compared to when using the MT model. When using PDVC, our framework achieves an overall accuracy (\ie weighted recall) of 50.22, which is only 0.59 points lower than the accuracy achieved when classifying videos using the gold standard captioned events of ActivityNet Captions. Taking the category imbalance in our dataset into account, we observe that the achieved macro-averages of precision, recall and F1 scores are much lower as compared to their weighted-averages. 
\extended{For extended results in text classification refer to the Supplementary Materials, Section~\ref{sec:extended_results}.}

\begin{table}
\scriptsize
\centering
    \begin{tabular}{l|l||N{21pt}N{21pt}N{21pt}}
        \toprule
        \multicolumn{5}{c}{\textit{\textbf{Text Classification}}} \\
        \toprule
        Averaging & Captioned events input & Prec. & Rec. & F1 \\
        \midrule 
        Weighted & MT & 43.72 & 48.24 & 44.80 \\
         & PDVC & 45.75 & 50.22 & 46.96 \\
        \cmidrule{2-5}
         & ActivityNet Captions & 46.97 & 50.81 & 48.23 \\
        \midrule 
        Macro & MT & 27.51 & 30.45 & 28.08\\
         & PDVC & 34.31 & 32.30 & 30.88 \\
        \cmidrule{2-5}
         & ActivityNet Captions & 33.42 & 34.74 & 33.19\\
        \bottomrule
    \end{tabular}
    \caption[Experimental results for the classification of videos.]{Results for classification of video captions in the settings: (i) captioned events generated by MT, (ii) PDVC, and (iii) gold standard captioned events from ActivityNet Captions. 
    For MT and PDVC, $|dist(E)| = 10$ is used. 
    \extended{Reported are weighted and macro-averages of precision, recall, and F1 scores across all categories on our classification test set.}}
    \label{tab:results_classification}
\end{table}

\section{Discussion}
\label{sec:discussion}

\extended{In this work, we propose a framework for the automatic extraction of multiple semantic metadata types from videos, namely the depicted \textit{entities} and their
\textit{properties}, the observable \textit{relations} between entities, and the video
\textit{category}. For each metadata type, we present a method which extracts the corresponding semantic information either directly from the captioned events that the framework's DVC model generates for an input video, or the linguistic annotations produced by the framework's language parser for the captioned events.
With the results presented in Section~\ref{sec:results}, we now discuss  the capability of our framework to extract semantic metadata, if and how the quality of extracted metadata is affected by the amount of information that is generated by its DVC model, and whether there is an influence of the performance of the framework's DVC model to the quality of semantic metadata that is extracted from the captioned events it generates.}
    
\subsection{Key Results}
\label{sec:discussion/key_results}
The experiments show that our proposed framework is able to automatically generate multiple types of semantic metadata in a meaningful way. We have to keep in mind that in our framework semantic metadata is extracted only from captioned events that are automatically generated by its DVC model, \ie a model that is designed and trained for a different task. 
To extract higher quality semantic information for each semantic metadata type, a dedicated computer vision method could be used, such as video object detection and video visual relation tagging. Here, however, our framework trades quality for effectiveness as it only requires training for one computer vision model, the DVC model. We must also bear in mind that the variety of entities, entity-property pairs, and relations that occur in our gold standards, and that are extracted by our framework, is large, in particular when compared to related computer vision tasks. For example, the VidOR (Video Object Relation) dataset~\cite{VidOR1, VidOR2}, which is used to train models for visual relation detection from videos, contains annotations of 80 categories of objects and 50 categories of relation predicates. In our gold standard for relations, however, 2,493 different entities acted either as subject (905 entities) or object (2,299 entities), while 905 different verbs occur in the relations. 
Also, for video-level entity extraction, we observed that for higher entity frequency thresholds $f$, the framework's precision decreases only slightly, while recall performance improves greatly. 
This observation is not surprising, as the number of different entities in our gold standard for entities, which is based on ActivityNet Captions, is large, and many entities occur only a few times\extended{ (see Table~\ref{tab:statistics_entity_frequency_threshold})}. 
This leads to the conclusion that the DVC models are only able to learn entities when there is enough training data, \ie the models see them sufficiently enough during training. 
\extended{Out of the semantic metadata types entities, properties, and relations, our framework achieves its best results in entity extraction.} 
Regarding the video classification results (based on the generated captions), our framework could not reliably predict the video category.
This is because the captioned events generated by the DVC models do not contain sufficiently specific semantic information for videos of different categories. 
Here, a visual-based video classifier is preferred over a purely text-based approach\extended{ based on video captions}.

\extended{We introduced the parameters $|E|$ and $|dist(E)|$, the number of (distinct) captioned events that are generated, to be able to control the amount of information that the DVC model provides for the semantic metadata extraction methods. 
The dense video captioning models MT and PDVC internally calculate confidence scores for their generated captioned events. If the number of considered captioned events is limited, then only the generated captioned events with the best confidence scores are used. The experiments show that the confidence scores serve their purpose. As expected, precision of predictions decreases for increasing $|E|$ and $|dist(E)|$ with only few exceptions. On the other hand, higher recall values are achieved when increasing $|E|$ and $|dist(E)|$. This applies to all, event localization and dense video captioning, as well as entity, property, and relation extraction. We assume the F1 score to be the metric that best measures the overall quality of extracted semantic metadata. For all extracted semantic metadata types, Table~\ref{tab:best_f1_performances} lists the combination of the DVC model and the number of its generated captioned events our framework uses to achieve best performance with respect to F1 score. We observe that usually for 10 generated captioned events the highest F1 scores are achieved, except for property extraction, where this is the case for $|dist(E)| = 25$. 
This indicates that the quality of semantic metadata extracted by our framework degrades when more than 10 captioned events are generated by its DVC model.}

\extended{
\begin{table}
\centering
\scriptsize
    \begin{tabular}{l||c|c}
         \toprule
        \multicolumn{1}{c||}{Semantic Metadata Type} & \multicolumn{1}{c|}{DVC Model} & \multicolumn{1}{c}{$|E$| \ / \ $|dist(E)|$}\\
        \midrule
        Entities (video-level)  & PDVC & $10$ \\
        Entities (event-level)  & PDVC & $10$ \\
        Properties  & MT & $25$ \\
        Relations (video-level)  & PDVC & $10$ \\
        Relations (event-level)  & PDVC & $10$ \\
        Category  & PDVC & $10$ \\
        \bottomrule
    \end{tabular}
    \caption[Parameter combinations for best F1 score performances of our framework.]{Combinations of DVC model and number of generated captioned events that result in the highest F1 scores of our framework for different semantic metadata types.}
    \label{tab:best_f1_performances}
\end{table}
}

\extended{Finally, we analyze whether there are correlations between the performance of the DVC model that is used by our framework and the quality of the extracted semantic metadata. Revisiting Table~\ref{tab:best_f1_performances}, we observe that our framework achieves highest F1 scores for all semantic metadata types when employing the PDVC model, except for property extraction. Looking at the dense video captioning results of MT and PDVC with respect to $|E|$ and $|dist(E)|$ (see Table~\ref{tab:results_dvc_per_E}), we also observe that PDVC usually outperforms MT regarding the METEOR score when generating distinct captioned events. 
For non-distinct captioned events, this is only the case for $|E| = 10$. 
For higher $|E|$, the achieved METEOR score of PDVC decreases more sharply as compared to the MT model due to declining precision for event localization. 
We made the same observations for event-level extraction of entities and relations.}

\subsection{Threats to Validity \shortorextended{and Future Work}{}}
\label{sec:discussion/threads}
We generated a gold standard using the captioned events of ActivityNet Captions validation videos. As described in Section~\ref{sec:experiments/datasets}, we use our proposed entity, property, and relation extraction methods on the processed captioned events, and treat the results as gold standards for semantic metadata. 
This requires that the extraction methods work sufficiently well, \ie they are able to extract semantic metadata using the linguistic annotations provided by the framework's language parser. 
In order to validate this hypothesis, we annotated a subset of $25$ ActivityNet Captions videos with the video-level entities, entity-property pairs, and video-level relations that we expect the methods to extract from captioned events. 
In total, we made annotations of 110 captioned events in these $25$ videos.
Table~\ref{tab:dataset_evaluation} shows the precision and recall performances of our entity, property, and relation extraction methods on these manually annotated videos. The results show that our entity and property extraction methods are certainly reliable. In some cases, however, our entity extraction method wrongly determines entities. For example, for the sentence \textit{"the camera pans around the field"}, spaCy wrongly classifies \textit{camera}, \textit{pans} and \textit{field} as nouns, while \textit{pans} actually acts as verb in the sentence. Since \textit{pan} is listed in WordNet as a noun, \textit{pans} is still determined as an entity. The relation extraction method is able to determine only around 70\% of the relations\extended{~that we annotated}. This is due to the complexity of relation extraction\extended{, which relies strongly on the dependency parse that the language parser of our framework produces}. 
\extended{For more complex sentences, we observed that the language parser produces syntactic relations between tokens with wrong labels, thus leading the relation extraction method to determine either no relations, or incorrect relations.}

\extended{It becomes clear that the quality of extracted semantic metadata depends heavily on the quality of the linguistic annotations that the language parser provides, \ie the POS tags, dependency labels, and coreference clusters. The quality of linguistic annotations, again, depends on whether the sentences at hand are grammatically correct or not. This means that the DVC model used by our framework needs to generate proper sentences with respect to grammar. Only then the language parser is able to generate linguistic annotations of high quality, ensuring the entity, property, and relation extraction methods to extract semantic information from linguistic annotations.}

\begin{table}[th]
\scriptsize
\centering
    \begin{tabular}{l|N{22pt}N{22pt}}
         \toprule
        \multicolumn{1}{c|}{Semantic Metadata Type} & \multicolumn{1}{c}{Prec.} & \multicolumn{1}{c}{Rec.}\\
        \midrule
        Entities (video-level)  & 94.21 & 98.39 \\ 
        Properties              & 92.98 & 91.38 \\ 
        Relations (video-level) & 78.36 & 70.62 \\ 
        \bottomrule
    \end{tabular}
    \caption{Evaluation of the entity, property, and relation extraction methods on captioned events from 25 manually annotated videos.}
    \label{tab:dataset_evaluation}
\end{table}

\extended{In Section~\ref{sec:method_entity_extraction}, we declared that, \eg an object is only an entity when it can be seen in the video. In our framework, we extract entities from the captioned events generated by a DVC model. However, the entities that appear in the sentence descriptions of the captioned events are not necessarily visible in the video (\eg \textit{camera} in \textit{"A man is speaking to the camera"}). 
To analyze how many entities that are extracted by our framework for ActivityNet Captions videos are actually not visible in the video, we used ActivityNet-Entities\footnote{ActivityNet-Entities: \url{https://github.com/facebookresearch/ActivityNet-Entities}.}~\cite{ActivityNetEntities}, a dataset based on ActivityNet Captions. ActivityNet-Entities augments 12,469 videos from the ActivityNet Caption's train and validation sets with bounding box annotations, each grounding a noun phrase (NP). In particular, NPs are not annotated when they are abstract or not presented in the scene. By comparing the entities that appear in ActivityNet-Entities with those included in our generated gold standard for entities, we can estimate how many extracted entities are actually not visible in the input video. Predicting the video-level entities extracted from NPs from ActivityNet-Entities with the video-level entities extracted from captioned events from ActivityNet Captions results in a precision of 64.97 and recall of 98.17. 
Therefore, up to two thirds of the entities in our gold standards for entities are actually abstract or not visible (\eg \textit{camera}, \textit{front}, \textit{end}, and \textit{game}). Furthermore, when extracted from a captioned event, an event-level entity has the same temporal segment as the captioned event. Therefore, it is assumed that the entity is visible for the entire duration of the temporal segment, even though the entity may actually be visible for only a part of it (\eg \textit{glass} in \textit{"the man puts a glass into a cupboard, then closes it and walks out of the room"}). Finally, it is also likely that many entities, properties, and relations that appear in a video are not captured by our framework. This is because the semantic information may not be relevant for a video, and therefore is not included in a caption event generated by the DVC model.}

\extended{Recall that when implementing and training the DVC models MT~\cite{MT} and PDVC~\cite{PDVC}, we changed few parameters compared to their original works (see Section \ref{sec:experiments/hyperparameters}). To ensure consistency, we evaluated both models in event localization and dense video captioning according to their original evaluation setup. 
Our trained models achieved similar, and for dense video captioning even better performance (see Supplementary Materials, Section~\ref{sec:extended_results}).}

\extended{
\subsection{Generalizability}
\label{sec:discussion/generalizability}

We stated that our framework is not able to differentiate whether an extracted entity is actually visible in a video or not. Since the gold standard captioned events of ActivityNet Captions also contain non-visible entities, the DVC model reproduces this characteristic, leading our framework to extract semantic metadata that is abstract or not presented in the video. 
In general, the quality of extracted semantic metadata depends on the quality of semantic information that the framework's DVC model generates, which again depends on the quality or characteristics of the dataset on which the DVC model is trained on. 
YouCook2~\cite{YouCook2}, \eg is a task-oriented video dataset in the domain of cooking. In this case, the annotated captioned events for videos are cooking instructions. The proportion of actually observable semantic information that YouCook2 annotations contain is much higher as compared to ActivityNet Captions annotations. Consequently, our framework would have the same characteristic when trained and used on YouCook2. Ultimately, this means that our framework produces semantic metadata of highest quality, when its video description model is trained with captioned events that explicitly only describe the observable entities, their properties, and how the entities relate to each other.

Our proposed framework extracts semantic metadata from videos by first generating semantic information in textual form using DVC, and then applying NLP methods. We decided to use a DVC model for generating semantic information as this allows us to associate entities and relations with temporal segments, \ie to generate event-level semantic metadata. 
When event-level information is not required, the DVC model can be easily replaced in our framework with other video description methods. 
For example, Video Paragraph Generation models~\cite{VideoParagraphGeneration, VideoParagraphGeneration2} automatically generate multiple natural language sentences that provide a narrative of a video clip, without having to localize individual events~\cite{SurveyVideoCaptioning1}. Furthermore, our framework can be used to extract semantic metadata from other multimedia types, \eg from images. Here, the framework's model for generating semantic information in textual form could be an Image Paragraph Captioning model~\cite{ImageParagraphCaptioning1, ImageParagraphCaptioning2}, which generate a detailed, multi-sentence description of the content of an image.

We conclude that our framework is easily adaptable because of its modularity: multiple configurations of our framework can be implemented, each serving its specific purpose. Different DVC models can be trained for different video domains (\eg cooking, animals) to improve the quality of generated semantic information. Further video description methods can be used in our framework, allowing the extraction of only video-level semantic metadata. The framework's text classification method, and entity, property, and relation extraction methods can be modified or replaced, and other semantic metadata extraction methods can easily be added. Lastly, by using an appropriate description model that generates semantic information in textual form, our framework can be adapted to extract semantic metadata from other multimedia types such as images.}

\extended{\subsection{Future Work}}
\label{sec:discussion/future_work}

\extended{As the capabilities of our proposed framework are not yet exhausted, our work enables several paths for future work. The modular structure of our framework allows to replace its DVC model, the language parser, and also the text classification method at will. Iashin and Rahtu~\cite{Bi-modal_DVC} argue that handling of multiple modalities is under-explored in the computer vision community. To this end, they presented a bi-modal transformer which uses both, visual features and the audio track, and achieved state-of-the-art performance for dense video captioning. Perhaps our framework produces higher quality semantic metadata when employing a DVC model that also uses the audio track.}
\extended{More generally, it would be interesting to see how well our framework performs for further DVC models, and different language parser. The training of the framework's DVC model could be improved for the purpose of semantic metadata extraction by putting more attention to, \eg entities and which properties they have. Also, instead of the spaCy library, in our framework the widely used Stanford CoreNLP toolkit~\cite{CoreNLP} could be used to generate linguistic annotations from text. However, this would involve minor changes to the entity, property, and relation extraction methods as, unlike spaCy, Stanford CoreNLP uses the dependency labels of Universal Dependencies.}

\extended{As mentioned above, in our framework the DVC model can be replaced easily with other multimedia description methods. 
In the domain of videos, it would be interesting to see how well our framework is able to extract video-level semantic metadata when using a video paragraph generation model instead of a DVC model~\cite{VideoParagraphGeneration, VideoParagraphGeneration2}. 
As opposed to video paragraph generation, DVC models generate sentences that are not necessarily coherent. 
Therefore, when in our framework text is generated with a video paragraph generation model, the language parser could be able to produce linguistic annotations of higher quality, resulting in improved quality of extracted semantic metadata.}

\extended{The entities that our framework extracts are generic. For example, when in a video \textit{Lionel Messi} is seen standing in the \textit{Camp Nou}, our framework would generate the entities \textit{man} and \textit{stadium}. To enable extraction of named entities, the framework's video description model could be supported by models that are able to, \eg identify persons, such as face recognition models~\cite{FaceRecognition}. This would also allow the language parser to use Named Entity Recognition~\cite{NER}, \ie the tagging of named entities in text with their corresponding type like \texttt{person} or \texttt{location}, potentially improving the quality of the linguistic annotations it produces.}

\extended{Finally, we considered the semantic metadata types entities, properties, relations, and the video category. However, depending on the application, one might prefer a more detailed distinction of entities, \eg between persons, objects, and locations, rather than just entities. Furthermore, other types of semantic metadata may be of interest, \eg emotions. Our framework can easily be extended by further metadata extraction methods. It would be interesting to see where the limits of the framework's video description model are in terms of what amount of useful semantic information it can generate for semantic metadata extraction.}

\section{Conclusion}
We presented a framework for metadata extraction of various types from generated video captions.
\extended{Our framework combines dense video captioning with natural language parsing, open information extraction, and text classification. }
The metadata quality mainly depends on two factors:
The event localization and video captioning performance of the dense video captioning model, and the number of captioned events forwarded from the dense video captioning model to the semantic metadata extraction methods.
This opens the path for future research on integrated models for semantic metadata extraction.

\paragraph{Acknowledgments}~This paper is the result of Johannes' Bachelor's thesis supervised by the two co-authors.
A short version of the work is published at CD-MAKE 2023 held at the University of Sannio in Benevento, Italy.
This long version of the paper is essentially the BSc thesis. 
We have explicitly documented the cut-out parts in blue color in the preprint version 2, which is directly available here: 
\url{https://arxiv.org/pdf/2211.02982v2.pdf}.

\vspace{3mm}

The concept behind the writing of theses as scientific papers is documented here:
\url{https://github.com/data-science-and-big-data-analytics/teaching-examples/blob/main/Scherp-TdL21-vortrag.pdf}.

\bibliographystyle{splncs04}
\bibliography{library-short}

\extended{
\clearpage

\addtocontents{toc}{\protect\setcounter{tocdepth}{1}} 

\appendix

\section*{Appendix}

\input{supplementary_materials/01_tags_and_dependency_labels}
\input{supplementary_materials/02_extended_results}
}

\end{document}

%% file: supplementary_materials/01_tags_and_dependency_labels.tex
\section{Tags and Dependency Labels}
\label{sec:tags_and_dependencies}
In this work, we use the Universal Dependencies POS tags and ClearNLP dependency labels when explaining how the framework's semantic metadata extraction methods use linguistic annotations produced by the language parser (see Sec.~\ref{sec:method}). Table~\ref{tab:universal_dependencies_tags} and Table~\ref{tab:clear_nlp_dependency_labels} list and name the POS tags and dependency labels mentioned in this work. 

\begin{table}[h]
    \begin{tabular}{l|l}
        \toprule
        POS Tag & Description\\
        \midrule
        \texttt{NOUN} & Noun \\ 
        \texttt{PRON} & Pronoun \\ 
        \texttt{PROPN} & Proper Noun \\ 
        \texttt{VERB} & Verb  \\ 
        \texttt{AUX} & Auxiliary verb \\ 
        \texttt{ADV} & Adverb \\ 
        \texttt{ADP} & Adposition \\ 
        \texttt{ADJ} & Adjective \\ 
        \bottomrule
    \end{tabular}
    \caption[List of Universal Dependencies part-of-speech Tags.]{List of the Universal Dependencies \cite{UniversalDependencies} part-of-speech Tags used in this work. For a complete list refer to the Universal Dependencies web page\protect\footnotemark.}
    \label{tab:universal_dependencies_tags}
\end{table}

\footnotetext{Universal Dependencies POS tags: \url{https://universaldependencies.org/u/pos/}.}

\begin{table*}
\small
    \begin{tabular}{p{2cm}|p{2.5cm}|p{8cm}}
        \toprule
        Dependency Label & Description & Explanation by ClearNLP \\
        \midrule
        \texttt{nsubj} & Nominal Subject             & Non-clausal constituent in the subject position of an active verb. \\
        \texttt{nsubjpass} & Nominal Subject (pass.) & Non-clausal constituent in the subject position of a passive verb. \\

        \texttt{auxpass} & Auxiliary (passive)       & Auxiliary verb that modifies a passive verb. \\
        \texttt{agent} & Agent                       & Complement of a passive verb that is the surface subject of its active form. \\
        \texttt{acl} & Clausal Modifier              & Either an infinitival modifier, i.e., a phrase that modifies the head of a noun phrase, \\
         &                                           & or a participial modifier, i.e., phrase whose head is a verb in a participial form \\
         & & that modifies the head of a noun phrase, or a complement.\\
         
        \texttt{dobj} & Direct Object                & Noun phrase that is the accusative object of a (di)transitive verb. \\
        \texttt{pobj}  & Object of Preposition       & Noun phrase that modifies the head of a prepositional phrase, which is usually a \\
        &                                            & preposition, but can be a verb in a participial form.\\

        \texttt{prt} & Particle                      & Preposition in a phrasal verb that forms a verb-particle construction. \\
        \texttt{prep} & Prepositional Modifier       & Any prepositional phrase that modifies the meaning of its head. \\
        \texttt{amod} & Adjectival Modifier          & Adjective or an adjective phrase that modifies the meaning of another word.\\
        \texttt{advmod}                              & Adverbial Modifier & Adverb or an adverb phrase that modifies the meaning of another word.\\
        
        \texttt{compound} & Compound                 & Either a noun modifying the head of noun phrase, a number modifying the head \\
         & & of quantifier phrase, or a hyphenated word.\\
        \texttt{conj} & Conjunct                     & Dependent of the leftmost conjunct in coordination. \\
        \bottomrule
    \end{tabular}
    \caption[List of ClearNLP dependency labels.]{List of the ClearNLP dependency labels used in this work. For a complete list refer to the ClearNLP Guidelines\protect\footnotemark.}
    \label{tab:clear_nlp_dependency_labels}
\end{table*}

\footnotetext{ClearNLP dependency labels: \url{https://github.com/clir/clearnlp-guidelines/blob/master/md/specifications/dependency_labels.md}.}

%% file: supplementary_materials/02_extended_results.tex
\section{Extended Results}
\label{sec:extended_results}
The following sections contain extended experimental results. 
In particular, we provide the entities, entity-property pairs, and relations that are most frequently extracted by our framework for ActivityNet Captions videos.
We present further experimental results for dense video captioning and video category extraction with our framework.
Finally, we evaluate whether the performance of the framework's DVC model for a particular video category relates with the framework's classification performance on the same category.

\subsection{Frequently Extracted Semantic Metadata}
Table~\ref{tab:statistics_frequent_semantic_metadata} shows the entities, entity-property pairs, and relations that are most frequently extracted by our framework for ActivityNet Captions videos. Recall that the frequency of a video-level semantic metadata instance is the number of different videos in which it occurs. For each, video-level entities (Table~\ref{tab:statistics_frequent_entities}), entity-property pairs (Table~\ref{tab:statistics_frequent_properties}), and video-level relations (Table~\ref{tab:statistics_frequent_relations}), we list the instances that occur most frequently in our corresponding generated gold standard, and the instances that are most frequently extracted by our framework for ActivityNet Captions validation videos when using MT \cite{MT} or PDVC \cite{PDVC} as dense video captioning model. For entities, we also list the gold standard entities whose names are compound nouns that were detected with our entity extraction method (see Section~\ref{sec:method_entity_extraction}).

\subsection{Dense Video Captioning}
When implementing and training the dense video captioning models MT and PDVC, we changed few parameters compared to their original works. To ensure consistency, we evaluate both models in event localization and dense video captioning according to their original evaluation setup. Therefore, we set $|E|=3$ for PDVC, while for MT the number of generated captioned events is not limited, and evaluate both models on the entire ActivityNet Captions validation set, using the official evaluation toolkits provided by ActivityNet Captions Challenge 2017 and 2018. Compared to the results in their original works, for MT and PDVC we achieved similar, and for dense video captioning even better, performances (Table~\ref{tab:results_dvc}). Since for the MT model the number of generated captioned events is not limited, it generates 210 captioned events per video on average. This explains its high recall and low precision regarding event localization performance compared to PDVC, which results in worse dense video captioning performance (using the 2018 evaluation toolkit). For our purpose, we think it is fairer to evaluate the DVC models as we did in Section~\ref{sec:results/dense_video_captioning}, i.e., to specify the number of captioned events that are maximally generated by PDVC and MT.

\subsection{Text Classification}
\begin{figure*}
    \centering
    \includegraphics[width=1.00\textwidth]{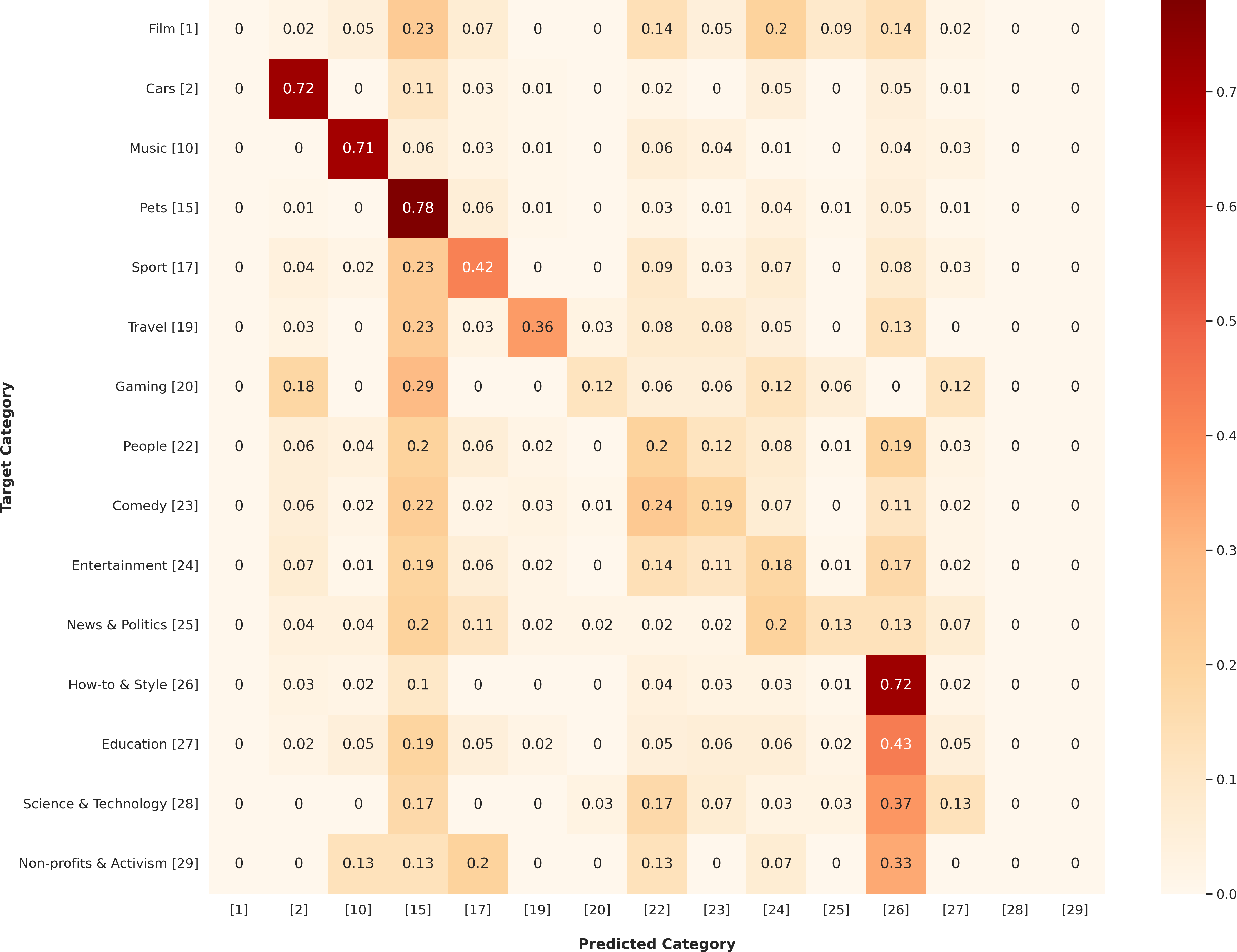}
    \caption[Confusion matrix for the classification method when trained with captioned events generated by MT.]{Confusion matrix for the classification method of our framework when trained with captioned events generated by MT.}
    \label{fig:confusion_matrix_MT}
\end{figure*}

\begin{figure*}
    \centering
    \includegraphics[width=1.00\textwidth]{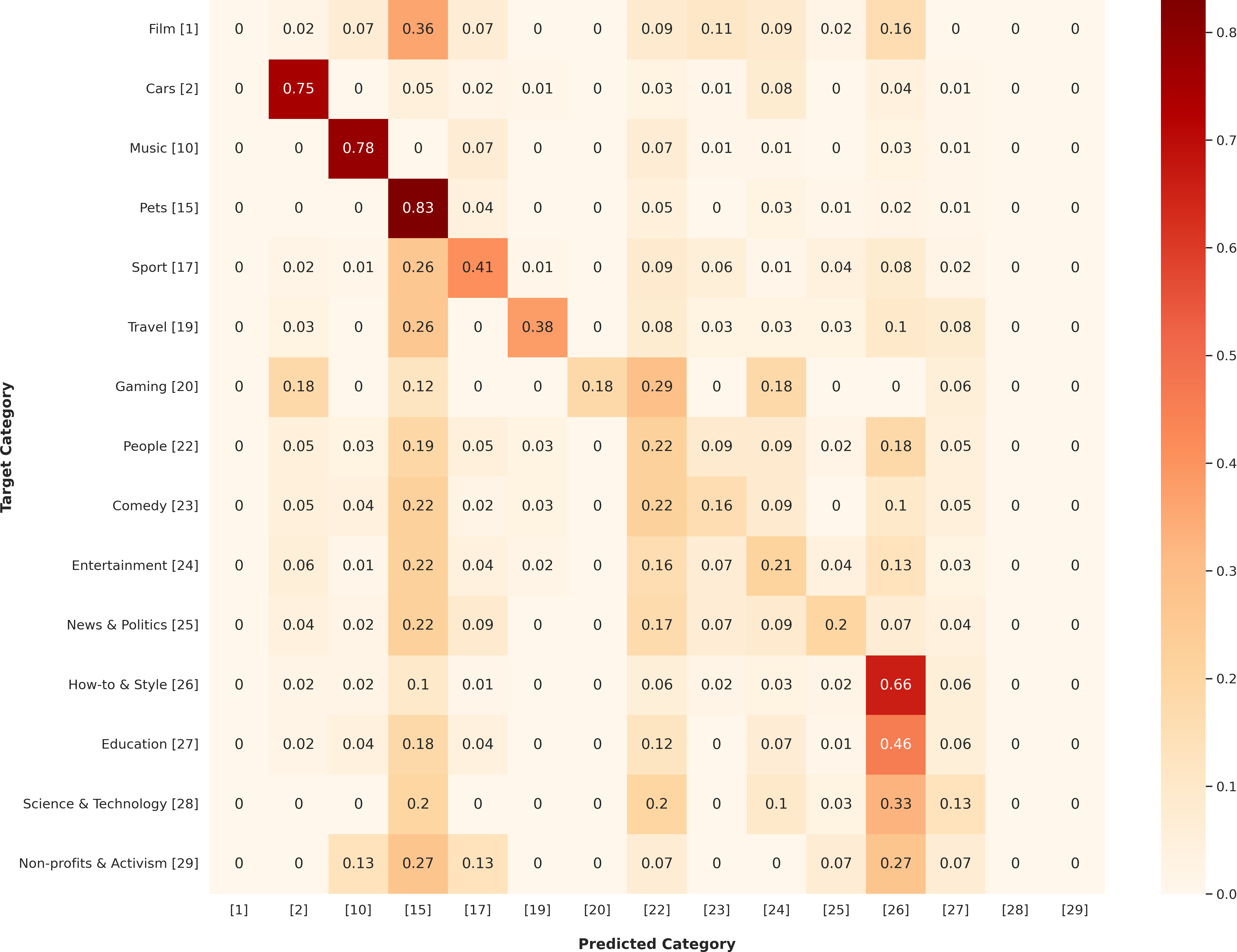}
    \caption[Confusion matrix for the classification method when trained with captioned events generated by PDVC.]{Confusion matrix for the classification method of our framework when trained with captioned events generated by PDVC.}
    \label{fig:confusion_matrix_PDVC}
\end{figure*}

\begin{figure*}
    \centering
    \includegraphics[width=1.00\textwidth]{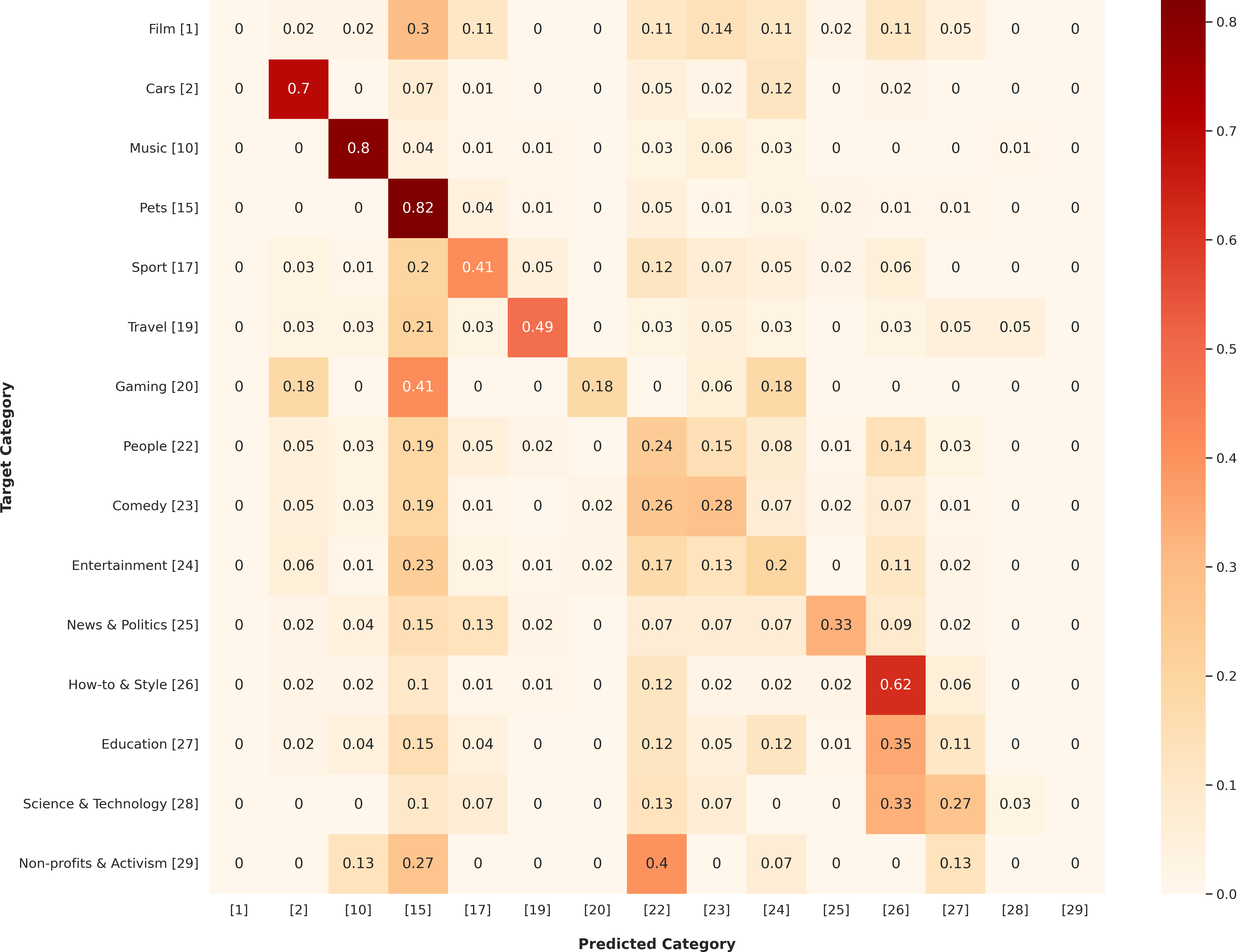}
    \caption[Confusion matrix for the classification method when trained with captioned events from ActivityNet Captions.]{Confusion matrix for the classification method of our framework when trained with gold standard captioned events from ActivityNet Captions.}
    \label{fig:confusion_matrix_ActivityNet}
\end{figure*}

Revisiting Table~\ref{tab:results_classification}, it shows the weighted and macro-averages of precision, recall, and F1 score performances of our framework’s
text classification method, when trained and evaluated in three different settings: using the captioned events generated by the MT and PDVC model, respectively, and using the captioned events provided by ActivityNet Captions. Taking the category imbalance in our classification dataset into account, we see that the achieved macro-averages of precision, recall, and F1 scores are much lower as compared to their weighted-averages. Looking at the classification performance per video category when using the PDVC model (see Table~\ref{tab:results_classification_PDVC}), we can observe that the highest F1 scores are achieved for the categories Sport, Music, and Pets \& Animals. However, when looking at the corresponding confusion matrix (see Figure~\ref{fig:confusion_matrix_PDVC}), we see that, even though we use class weights during training to overcome class imbalance, the categories Film \& Animation, Science \& Technology, and Non-profits \& Activism are not even predicted once. This explains the poor performance of our framework for macro precision, recall, and F1 score. 
Similar observations can be made when captioned events are used for classifications that are generated by the MT model (see Table~\ref{tab:results_classification_MT} and Figure~\ref{fig:confusion_matrix_MT}), or captioned events from ActivityNet Captions (see Table~\ref{tab:results_classification_ActivityNet} and Figure~\ref{fig:confusion_matrix_ActivityNet}).

\subsection{Analyzing DVC and Text Classification Performances per Video Category}

We showed that the quality of semantic metadata extracted by our framework depends heavily on the event localization and dense video captioning performance of the framework's DVC model. Finally, we now want to evaluate whether a similar statement applies to the classification of videos with our framework, i.e., whether the performance of the framework's DVC model for a particular video category relates with the framework's classification performance on the same category. Note that if a DVC model performs well on a video category, this does not logically translate into better classification performance of our framework. For DVC, a model performs well when it is able to describe the events of a video with appropriate natural language sentences. Theoretically, the gold standard annotations for videos of two different categories could contain similar words that, nevertheless, describe the videos well. For successful video category prediction, however, videos of different categories need to be distinguishable through unique features. These can be, e.g., unique entities that occur only for videos of a specific category.

Table~\ref{tab:results_dvc_per_category} shows the dense video captioning performances of MT and PDVC with respect to the video category. For both models, we set $|dist(E)| = 10$, the same number of distinct captioned events that is used for the classification of videos. Both models perform best on videos from the categories Pets \& Animals, Music, and Gaming. 
Regarding the categories Film \& Animation and Non-profits \& Activism, the DVCs have some of the lowest METEOR scores. 
The PDVC model achieves its second lowest METEOR score for How-to \& Style videos, although $1,661$ videos are available for this category, while How-to \& Style is among the top 5 categories for the MT model. 

However, for Gaming videos only a F1 score of 15.38 is achieved for text classification. For the categories Film \& Animation, Non-profits \& Activism, and News \& Politics the MT model achieves its worst METEOR performances. For these categories, also poor performances are achieved for text classification, with a F1 score of 16.22 for News \& Politics, and 0.00 else. Thus, if the MT model performs well (or poorly) for a certain video category, our framework seems to achieve better (or worse) classification performance in terms of F1 score for this particular video category. Nevertheless, there are exceptions, e.g., MT achieves relatively poor DVC performance for Sport videos, while the best F1 score performance for text classification with captioned events generated by MT is achieved for the Sports category. Similar observations can be made when classifying videos with the captioned events generated by the PDVC model (see Table~\ref{tab:results_classification_ActivityNet}).


\begin{table*}
\centering

\begin{minipage}{\textwidth}
    \begin{subtable}[h]{\textwidth}
    \centering
    \scriptsize
    \begin{tabular}{l|r||l|r||l|r||l|r}
        \toprule
        \multicolumn{8}{c}{\textit{Video-level Entity Extraction}} \\
        \toprule
         \multicolumn{2}{c||}{} &  \multicolumn{2}{c||}{(only Compound Nouns)} & \multicolumn{2}{c||}{Framework} & \multicolumn{2}{c}{Framework} \\
         \multicolumn{2}{c||}{Gold Standard} & \multicolumn{2}{c||}{Gold Standard} & \multicolumn{2}{c||}{with MT} & \multicolumn{2}{c}{with PDVC}\\
        Entity & Videos & Entity & Videos & Entity & Videos & Entity & Videos\\
        \midrule
        man & 2,450 & diving\_board & 31 & camera & 3,022 & man & 3,477 \\
        people & 1,262 & living\_room & 28 & man & 2,956 & camera & 3,124 \\
        camera & 1,144 & t-shirt & 26 & people & 1,506 & people & 1,300 \\
        woman & 1,082 & bumper\_car & 25 & woman & 1,016 & woman & 1,083 \\
        person & 679 & lawn\_mower & 25 & group & 772 & group & 799 \\
        hand & 574 & christmas\_tree & 25 & person & 618 & person & 757 \\
        girl & 517 & balance\_beam & 22 & pan & 545 & room & 683 \\
        front & 498 & tennis\_racket & 22 & movement & 544 & game & 615 \\
        group & 458 & parking\_lot & 20 & game & 476 & field & 399 \\
        water & 419 & jump\_rope & 19 & angle & 463 & chair & 377 \\
        \bottomrule
    \end{tabular}
    \caption{Entities that occur most frequently in our gold standard, and the entities that are most frequently extracted by our framework for different DVC models. Furthermore, we list the most frequent gold standard entities that consist of a compound noun.}
    \label{tab:statistics_frequent_entities}
    \end{subtable}
\end{minipage}

\vspace{2mm}

\begin{minipage}{\textwidth}
    \begin{subtable}[h]{\textwidth}
    \centering
    \scriptsize
    \begin{tabular}{l|r||l|r||l|r}
        \toprule
        \multicolumn{6}{c}{\textit{Property Extraction}} \\
        \toprule
         \multicolumn{2}{c||}{Gold Standard} & \multicolumn{2}{c||}{Framework with MT} & \multicolumn{2}{c}{Framework with PDVC}\\
        Entity [Property] & Videos & Entity [Property] & Videos & Entity [Property] & Videos\\
        \midrule
        group [large] & 126 & angle [several] & 463 & shot [several] & 344\\
        people [several] & 105 & shot [several] & 406 & group [large] & 306\\
        man [young] & 102 & group [large] & 387 & shirt [black] & 124\\
        shot [several] & 92 & object [various] & 132 & clip [several] & 98\\
        people [other] & 73 & clip [several] & 121 & girl [little] & 84\\
        girl [little] & 71 & field [large] & 94 & girl [young] & 71\\
        girl [young] & 69 & gymnasium [large] & 85 & boy [young] & 65\\
        clip [several] & 69 & people [several] & 64 & shirt [blue] & 58\\
        time [several] & 69 & girl [young] & 63 & field [large] & 53\\
        motion [slow] & 68 & crowd [large] & 38 & people [several] & 52\\
        \bottomrule
    \end{tabular}
    \caption{Entity-property pairs that occur most frequently in our gold standard, and the entity-property pairs that are most frequently extracted by our framework for different DVC models.}
    \label{tab:statistics_frequent_properties}
    \end{subtable}
\end{minipage}

\vspace{2mm}

\begin{minipage}{\textwidth}
    \begin{subtable}[h]{\textwidth}
    \centering
    \scriptsize
    \begin{tabular}{l|r||l|r||l|r}
        \toprule
        \multicolumn{6}{c}{\textit{Video-level Relation Extraction}} \\
        \toprule
         \multicolumn{2}{c||}{Gold Standard} & \multicolumn{2}{c||}{Framework with MT} & \multicolumn{2}{c}{Framework with PDVC}\\
        Relation & Videos & Relation & Videos & Relation & Videos\\
        \midrule
        (man, speak, [to], camera) & 137 & (man, speak, [to], camera) & 1,214 & (man, speak, [to], camera) & 1,684\\
        (woman, speak, [to], camera) & 92 & (man, lead, [into], man) & 587 & (man, lead, [into], man) & 1,273\\
        (man, talk, [to], camera) & 76 & (woman, speak, [to], camera) & 577 & (woman, speak, [to], camera) & 575\\
        (man, hit, [], ball) & 59 & (camera, capture, [], movement) & 488 & (man, stand, [in], room) & 312\\
        (man, stand, [in], front) & 51 & (camera, capture, [from], angle) & 463 & (man, play, [], game) & 312\\
        (man, throw, [], ball) & 39 & (camera, capture, [], man) & 226 & (man, end, [with], man) & 305\\
        (people, watch, [on], side) & 36 & (man, play, [], game) & 156 & (man, lead, [into], shot) & 250\\
        (man, play, [], drum) & 33 & (group, lead, [into], group) & 156 & (man, begin, [to], camera) & 219\\
        (woman, talk, [to], camera) & 31 & (camera, capture, [], people) & 153 & (woman, lead, [into], woman) & 219\\
        (man, hold, [in], hand) & 28 & (man, lead, [into], clip) & 152 & (people, play, [], game) & 214\\
        \bottomrule
    \end{tabular}
    \caption{Relations that occur most frequently in our gold standard, and the relations that are most frequently extracted by our framework for different DVC models.}
    \label{tab:statistics_frequent_relations}
    \end{subtable}
\end{minipage}

\caption[Statistics on most frequently extracted video-level semantic metadata by our framework.]{Statistics on video-level semantic metadata that is most frequently extracted by our framework when using MT and PDVC as dense video captioning model. For both, MT and PDVC, we limit the number of generated distinct captioned events \bm{$|dist(E)| = 3$}. For comparison, we list the most frequent entities, entity-property pairs and relations of our corresponding gold standards. All entities' names, entity-property pairs' names, and relations' names are lemmatized.}
\label{tab:statistics_frequent_semantic_metadata}

\end{table*}

\clearpage

\begin{table*}
    \scriptsize
    \centering
    \begin{tabular}{l|c||cc|ccc|ccc}
        \toprule
        \multicolumn{10}{c}{\textit{Event Localization \& Dense Video Captioning}} \\
        \toprule
         &  & 
        \multicolumn{1}{c}{avg.} & 
        \multicolumn{1}{c|}{avg.} & 
        \multicolumn{3}{c|}{2017 eval. toolkit} & 
        \multicolumn{3}{c}{2018 eval. toolkit} \\
        DVC Model & \multicolumn{1}{M{0.7cm}||}{$|E|$} &
        Recall & 
        Precision & 
        \multicolumn{1}{M{0.7cm}}{B@3} & 
        \multicolumn{1}{M{0.7cm}}{B@4} & 
        \multicolumn{1}{M{0.7cm}|}{M} & 
        \multicolumn{1}{M{0.7cm}}{B@3} & 
        \multicolumn{1}{M{0.7cm}}{B@4} & 
        \multicolumn{1}{M{0.7cm}}{M} \\
        \midrule
        MT & - & 91.48 & 32.11 & 5.47 & 2.63 & 10.21 & 2.65 & 1.30 & 5.19 \\
        PDVC & 3 & 50.19 & 57.87 & 4.65 & 2.32 & 9.12 & 4.14 & 2.07 & 8.20 \\
        \bottomrule
    \end{tabular}
    \caption[Event localization and dense video captioning results.]{Event localization and dense video captioning results on the ActivityNet Captions validation set. Both annotation files of the validation set are used for the evaluation. According to the setup in their original works, we set \bm{$|E| =3 $} for PDVC, while for MT the number of generated captioned events is not limited.\\}
    \label{tab:results_dvc}
\end{table*}

\begin{table*}
\centering

\begin{minipage}{.6\textwidth}
\centering
  \begin{subtable}[h]{\textwidth}
    \scriptsize
    \centering
    \begin{tabular}{r|l|r|ccc}
        \toprule
        \multicolumn{6}{c}{\textit{Dense Video Captioning with MT}} \\
        \toprule
        \multicolumn{2}{l|}{Category} & 
        \multicolumn{1}{c|}{} &
        \multicolumn{3}{c}{2018 eval. toolkit} \\
        \multicolumn{1}{c|}{ID} &
        \multicolumn{1}{c|}{Name} & 
        \multicolumn{1}{c|}{Videos} &
        \multicolumn{1}{M{0.57cm}}{B@3} & 
        \multicolumn{1}{M{0.57cm}}{B@4} & 
        \multicolumn{1}{M{0.57cm}}{M} \\
        \midrule
        15 & Pets \& Animals         & 342 & 2.82 & 1.27 & 6.88\\
        10 & Music                   & 637 & 3.12 & 1.54 & 6.83 \\
        20 & Gaming                  & 78 & 3.26 & 1.46 & 6.22\\
        19 & Travel \& Events        & 526 & 2.57 & 1.18 & 6.02\\
        26 & How-to \& Style         & 1,661 & 2.74 & 1.37 & 6.00\\
        22 & People \& Blogs         & 2,098 & 2.38 & 1.13 & 5.99\\
        23 & Comedy                  & 645 & 2.20 & 1.04 & 5.98\\
        24 & Entertainment           & 1,414 & 2.56 & 1.28 & 5.97\\
        27 & Education               & 414 & 2.37 & 1.03 & 5.84\\
        2 & Cars                     & 191 & 2.19 & 0.94 & 5.84\\
        17 & Sport                   & 3,920 & 2.47 & 1.24 & 5.83\\
        28 & Science \& Technology   & 146 & 2.29 & 0.95 & 5.70\\
        1 & Film \& Animation        & 213 & 1.74 & 0.64 & 5.46\\
        29 & Non-profits \& Activism & 68 & 2.25 & 1.02 & 5.42\\
        25 & News \& Politics        & 226 & 1.98 & 0.97 & 5.38\\
        \midrule
        \multicolumn{2}{r|}{weighted avg. across categories} & 12,579 & 2.51 & 1.22 & 5.98 \\
        \bottomrule
    \end{tabular}
    \caption{Dense video captioning performance of MT per category.}
    \label{tab:results_dvc_per_category_MT}
  \end{subtable}
\end{minipage}

\begin{minipage}{.6\textwidth}
  \begin{subtable}[h]{\textwidth}
    \scriptsize
    \centering
    \begin{tabular}{r|l|r|ccc}
        \toprule
        \multicolumn{6}{c}{\textit{Dense Video Captioning with PDVC}} \\
        \toprule
        \multicolumn{2}{l|}{Category} & 
        \multicolumn{1}{c|}{} &
        \multicolumn{3}{c}{2018 eval. toolkit} \\
        \multicolumn{1}{c|}{ID} &
        \multicolumn{1}{c|}{Name} & 
        \multicolumn{1}{c|}{Videos} &
        \multicolumn{1}{M{0.57cm}}{B@3} & 
        \multicolumn{1}{M{0.57cm}}{B@4} & 
        \multicolumn{1}{M{0.57cm}}{M} \\
        \midrule
        10 & Music                  & 637  & 4.25 & 2.31 & 7.40\\
        15 & Pets \& Animals        & 342  & 3.94 & 1.89 & 7.26\\
        20 & Gaming                 & 78   & 3.80 & 1.97 & 6.89\\
        2 & Cars                    & 191  & 3.13 & 1.45 & 6.86\\
        22 & People \& Blogs        & 2,098 & 3.46 & 1.71 & 6.72\\
        19 & Travel \& Events       & 526  & 3.38 & 1.68 & 6.69\\
        23 & Comedy                 & 645  & 3.26 & 1.57 & 6.67\\
        24 & Entertainment          & 1,414 & 3.30 & 1.69 & 6.61\\
        28 & Science \& Technology  & 146  & 3.20 & 1.57 & 6.45\\
        17 & Sport                  & 3,920 & 3.45 & 1.78 & 6.43\\
        25 & News \& Politics       & 226  & 3.50 & 1.83 & 6.36\\
        27 & Education              & 41   & 3.19 & 1.54 & 6.26\\
        1 & Film \& Animation       & 213  & 2.63 & 1.12 & 6.07\\
        26 & How-to \& Style        & 1,661 & 2.69 & 1.42 & 5.92\\
        29 & Non-profits \& Activism & 68  & 2.96 & 1.45 & 5.54\\
        \midrule
        \multicolumn{2}{r|}{weighted avg. across categories} & 12,579 & 3.44 & 1.75 & 6.57\\
        \bottomrule
    \end{tabular}
    \caption{Dense video captioning performance of PDVC per category.}
    \label{tab:results_dvc_per_category_PDVC}
  \end{subtable}
\end{minipage}

\caption[Dense video captioning results per video category.]{Dense video captioning results of MT (a) and PDVC (b) per video category on ActivityNet Captions. Here, we evaluate on the train and validation set, while using only the first of the two annotation files of the validation set. This is the same data that we use for the classification of videos. For both, MT and PDVC, we set \bm{$|dist(E)| = 10$}. The categories are sorted by the METEOR performances achieved on them.}
\label{tab:results_dvc_per_category}
\end{table*}

\clearpage
\begin{table*}
    \scriptsize
    \centering
    \begin{tabular}{r|l|r|ccc}
        \toprule
        \multicolumn{6}{c}{\textit{Text Classification}} \\
        \toprule        
        ID & Name & Videos & Prec. & Rec. & F1\\
        \midrule
        17 & Sport                    & 784  & 67.36 & 77.93 & 72.26 \\
        10 & Music                    & 128  & 53.18 & 71.88 & 61.13 \\
        15 & Pets \& Animals          & 69   & 52.69 & 71.01 & 60.49 \\
        26 & How-to \& Style          & 333  & 47.33 & 71.77 & 57.04 \\
        19 & Travel \& Events         & 106  & 28.85 & 42.45 & 34.35 \\
        2 & Cars \& Vehicles          & 39   & 31.82 & 35.90 & 33.73 \\
        22 & People \& Blogs          & 421  & 35.78 & 19.71 & 25.42 \\
        24 & Entertainment            & 284  & 28.65 & 17.96 & 22.08 \\
        23 & Comedy                   & 129  & 16.33 & 18.60 & 17.39 \\
        25 & News \& Politics         & 46   & 21.43 & 13.04 & 16.22 \\
        20 & Gaming                   & 17   & 22.22 & 11.76 & 15.38 \\
        27 & Education                & 84   &  7.02 &  4.76 &  5.67 \\
        1 & Film \& Animation         & 44   &  0.00 &  0.00 &  0.00 \\
        28 & Science \& Technology    & 30   &  0.00 &  0.00 &  0.00 \\
        29 & Non-profits \& Activism  & 15   &  0.00 &  0.00 &  0.00 \\
         \midrule
        \multicolumn{2}{r|}{weighted average} & 2,529 & 43.72 & 48.24 & 44.80 \\
        \multicolumn{2}{r|}{macro-average} & 2,529 & 27.51 & 30.45 & 28.08\\
        \bottomrule
    \end{tabular}
    \caption[Classification performance per video category when using captioned events generated by MT.]{Classification performance of our framework per video category when the text classifier is trained and evaluated with captioned events generated by MT. Categories are sorted according to the F1 score achieved for them.}
    \label{tab:results_classification_MT}
\end{table*}

\clearpage

\begin{table*}
    \centering
    \scriptsize
    \begin{tabular}{r|l|r|ccc}
        \toprule
        \multicolumn{6}{c}{\textit{Text Classification}} \\
        \toprule     
        ID & Name & Videos & Prec. & Rec. & F1\\
        \midrule
        17 & Sport                    & 784  & 68.00 & 82.65 & 74.61 \\
        10 & Music                    & 128  & 60.38 & 75.00 & 66.90 \\
        15 & Pets \& Animals          & 69   & 57.45 & 78.26 & 66.26 \\
        26 & How-to \& Style          & 333  & 50.00 & 66.37 & 57.03 \\
        2 & Cars \& Vehicles          & 39   & 37.50 & 38.46 & 37.97 \\
        19 & Travel \& Events         & 106  & 31.62 & 40.57 & 35.54 \\
        20 & Gaming                   & 17   & 100.0 & 17.65 & 30.00 \\
        22 & People \& Blogs          & 421  & 33.33 & 22.33 & 26.74 \\
        24 & Entertainment            & 284  & 34.08 & 21.48 & 26.35 \\
        25 & News \& Politics         & 46   & 16.36 & 19.57 & 17.82 \\
        23 & Comedy                   & 129  & 19.63 & 16.28 & 17.80 \\
        27 & Education                & 84   &  6.33 &  5.95 &  6.13 \\
        1 & Film \& Animation         & 44   &  0.00 &  0.00 &  0.00 \\
        28 & Science \& Technology    & 30   &  0.00 &  0.00 &  0.00 \\
        29 & Non-profits \& Activism  & 15   &  0.00 &  0.00 &  0.00 \\
         \midrule
        \multicolumn{2}{r|}{weighted average} & 2,529 & 45.75 & 50.22 & 46.96 \\
        \multicolumn{2}{r|}{macro-average} & 2,529 & 34.31 & 32.30 & 30.88 \\
        \bottomrule
    \end{tabular}
    \caption[Classification performance per video category when using captioned events generated by PDVC.]{Classification performance of our framework per video category when the text classifier is trained and evaluated with captioned events generated by PDVC. Categories are sorted according to the F1 score achieved for them.}
    \label{tab:results_classification_PDVC}
\end{table*}

\clearpage

\begin{table*}
    \centering
    \scriptsize
    \begin{tabular}{r|l|r|ccc}
        \toprule
        \multicolumn{6}{c}{\textit{Text Classification}} \\
        \toprule     
        ID & Name & Videos & Prec. & Rec. & F1\\
        \midrule
        17 & Sport                    & 784  & 68.76 & 82.27 & 74.91 \\
        15 & Pets \& Animals          & 69   & 59.78 & 79.71 & 68.32 \\
        10 & Music                    & 128  & 58.82 & 70.31 & 64.06 \\
        26 & How-to \& Style          & 333  & 55.91 & 62.46 & 59.01 \\
        2 & Cars \& Vehicles          & 39   & 43.18 & 48.72 & 45.78 \\
        19 & Travel \& Events         & 106  & 34.40 & 40.57 & 37.23 \\
        25 & News \& Politics         & 46   & 32.61 & 32.61 & 32.61 \\
        22 & People \& Blogs          & 421  & 32.70 & 24.47 & 27.99 \\
        24 & Entertainment            & 284  & 32.58 & 20.42 & 25.11 \\
        23 & Comedy                   & 129  & 20.22 & 27.91 & 23.45 \\
        20 & Gaming                   & 17   & 30.00 & 17.65 & 22.22 \\
        27 & Education                & 84   & 12.33 & 10.71 & 11.46 \\
        28 & Science \& Technology    & 30   & 20.00 &  3.33 &  5.71 \\
        1 & Film \& Animation         & 44   &  0.00 &  0.00 &  0.00 \\
        29 & Non-profits \& Activism  & 15   &  0.00 &  0.00 &  0.00 \\
         \midrule
        \multicolumn{2}{r|}{weighted average} & 2,529 & 46.97 & 50.81 & 48.23 \\
        \multicolumn{2}{r|}{macro-average} & 2,529 & 33.42 & 34.74 & 33.19\\
        \bottomrule
    \end{tabular}
    \caption[Classification performance per video category when using captioned events from ActivityNet Captions.]{Classification performance of our framework per video category when the text classifier is trained and evaluated with captioned events from ActivityNet Captions. Categories are sorted according to the F1 score achieved for them.}
    \label{tab:results_classification_ActivityNet}
\end{table*}

\clearpage